\documentclass[journal]{IEEEtran}
\usepackage{newtxtext}

\usepackage{hyperref}

\usepackage{booktabs}
\usepackage{multirow}
\usepackage{tablefootnote}
\usepackage{array}
\usepackage{authblk}
\usepackage{xspace}
\usepackage[table]{xcolor}
\usepackage{subfigure}
\usepackage{booktabs}

\usepackage{amsmath}
\usepackage{amssymb} 
\usepackage[para]{footmisc}

\ifCLASSINFOpdf
  \usepackage[pdftex]{graphicx}
\else
\fi
\usepackage{amsfonts, amsmath, amssymb}
\usepackage{pifont}
\usepackage{amsmath}

\usepackage[]{mdframed}
\usepackage{algpseudocode}

\usepackage{url}

\usepackage[most]{tcolorbox}

\newtcolorbox{insightbox}{
  colback=black!2,
  colframe=black!25,
  boxrule=0.4pt,
  sharp corners,
  left=2pt,
  right=2pt,
  top=1.5pt,
  bottom=1.5pt,
  before skip=3pt,
  after skip=2pt
}

\newcommand{\ie}{i.e.,\xspace}

\newcommand{\vct}[1]{\ensuremath{\boldsymbol{#1}}}

\newcommand{\set}[1]{\ensuremath{\mathcal{#1}}}

\newcommand{\myparagraphdot}[1]{\noindent\textbf{#1.}}
\newcommand{\myparagraph}[1]{\noindent\textbf{#1}}

\begin{document}
%
\title{Label-efficient Training Updates for Malware Detection over Time}
%
%
%
\author[1]{Luca Minnei}
\author[1]{Cristian Manca}
\author[1]{Giorgio Piras}
\author[1]{Angelo Sotgiu}
\author[1]{Maura Pintor,~\IEEEmembership{Member,~IEEE}}
\author[1,2]{Daniele Ghiani}
\author[1]{Davide Maiorca,~\IEEEmembership{Member,~IEEE}}
\author[1,3]{Giorgio Giacinto,~\IEEEmembership{Senior Member,~IEEE}}
\author[1,3]{Battista Biggio,~\IEEEmembership{Fellow,~IEEE}}

\affil[1]{Department of Electrical and Electronic Engineering,
University of Cagliari, Italy}
\affil[2]{Department of Computer, Control and Management Engineering, Sapienza University, Rome, Italy}
\affil[3]{CINI, Rome, Italy}

\renewcommand{\subsectionautorefname}{Sect.}
\renewcommand{\sectionautorefname}{Sect.}
\renewcommand{\figureautorefname}{Fig.}


\maketitle

\begin{abstract}
Machine Learning (ML)-based detectors are becoming essential to counter the proliferation of malware. However, common ML algorithms are not designed to cope with the dynamic nature of real-world settings, where both legitimate and malicious software evolve. This \textit{distribution drift} causes models trained under static assumptions to degrade over time unless they are continuously updated.
Regularly retraining these models, however, is expensive, since labeling new acquired data requires costly manual analysis by security experts.
To reduce labeling costs and address distribution drift in malware detection, prior work explored active learning (AL) and semi-supervised learning (SSL) techniques. Yet, existing studies (i)~are tightly coupled to specific detector architectures and restricted to a specific malware domain, resulting in non-uniform comparisons; 
and (ii)~lack a consistent methodology for analyzing the distribution drift, despite the critical sensitivity of the malware domain to temporal changes.
In this work, we bridge this gap by proposing a model-agnostic framework that evaluates an extensive set of AL and SSL techniques, isolated and combined, for Android and Windows malware detection. We show that these techniques, when combined, can reduce manual annotation costs by up to 90\% across both domains while achieving comparable detection performance to full-labeling retraining.
We also introduce a methodology for feature-level drift analysis that measures feature stability over time, showing its correlation with the detector performance.
Overall, our study provides a detailed understanding of how AL and SSL behave under distribution drift and how they can be successfully combined, offering practical insights for the design of effective detectors over time.

\end{abstract}


%
\IEEEpeerreviewmaketitle

\section{Introduction}\label{Sect:Intro}
Android and Windows dominate mobile and desktop ecosystems, making them prime targets for \textit{malware}, malicious software that can infiltrate and compromise devices.
Machine Learning (ML) algorithms enable automatic and effective malware detection~\cite{arp2014drebin, anderson2018ember}. However, such detectors are often evaluated under stationary conditions, with train and test data drawn from the same underlying distribution.
This fails to reflect real-world settings where both malware and benign samples continuously evolve, inducing significant shifts in data distributions. This is commonly referred to as \textit{distribution drift}, which leads to progressive detector performance decay when tested in the wild~\cite {pendlebury2019tesseract}.
A common countermeasure is to continually retrain detectors on updated data. 
However, labeling newly collected samples can be resource- and time-consuming, as it often requires expert inspection~\cite{joyce2023MOTIF}.
Therefore, traditional fully supervised training is unrealistic in real-world scenarios.
To reduce labeling costs, prior work explored \textit{active learning} (AL)~\cite{settles2009active}, which selects informative samples for labeling, and \textit{semi-supervised learning} (SSL)~\cite{vanEngelen2020}, which assigns pseudo-labels to unlabeled data.
While promising, both have several drawbacks when applied in isolation: AL is constrained by the high annotation cost, resulting in small annotation budgets that may not capture the rapidly changing data distribution, while SSL can easily reinforce existing biases and propagate errors when pseudo-labels are incorrect~\cite{chen2022debiased}.
These limitations motivate their combination: AL focuses the limited labeling budget on a few highly informative samples, while SSL adds many pseudo-labeled samples from already observed distributions.
Existing work combining AL and SSL for distribution drift in malware detection still suffers from several limitations.
First, many implementations are tightly coupled to specific ML algorithms~\cite{alam2024morphautomatedconceptdrift} or ad hoc pipelines~\cite{guo2024malosdf}, limiting their generalizability across different models and settings.
Second, evaluations are typically confined to a single platform or domain (e.g., Android or Windows)~\cite{chen2023continuous, liu2025ldcdroid,minnei2025experimental} and with different baselines and setups, leaving it unclear whether their performance can be replicated across different environments.
Finally, while framed as distribution-drift remedies, prior work provides limited analysis and insights on drift and its causes.
In this work, we address these issues by proposing a model-agnostic framework combining AL and SSL for label-efficient retraining of malware detectors under distribution drift. 
We conduct a systematic evaluation of 8 AL and 2 SSL techniques, analyzing their individual and combined effects under distribution drift across Android and Windows malware datasets. 
To interpret the results, we introduce a feature-level stability analysis that captures how discriminative features evolve over time. 
Our findings show that combining AL and SSL achieves performance comparable to full retraining while reducing labeling costs by up to $90\%$. The proposed drift analysis further explains these results, highlighting why different retraining strategies exhibit different performance over time.
\section{Background}
This section provides essential background on ML for malware detection and on AL and SSL methodologies. 

\subsection{Machine Learning for Malware Detection}\label{Sect:ML}
ML techniques have been increasingly adopted for malware detection~\cite{MUZAFFAR2022androidMLSurvey, MANIRIHO2024windowsMLSurvey}. In this work, we focus on methods that extract features by running static analysis on input programs (e.g., Android apps or Windows PE files) and converting the output into a suitable numerical representation.
Each sample, associated with a label $y\in\{0,1\}$ (respectively, benign and malware), is mapped to a $d$-dimensional feature vector $\vct x \in \set X \subseteq \mathbb{R}^{d}$.
The ML detector $f$ is then trained on $(\vct{x},y)$ pairs to perform malware classification.
At inference time, the classifier takes as input a new sample $\vct x$ and outputs a confidence score $f(\vct x) \in [0, 1]$, representing the estimated probability that it belongs to the malware class; a predicted label $\hat y$ is then obtained by applying a decision rule $\hat y (\vct x) =\mathbb{I}[f(\vct x)>\tau]$, where $\mathbb{I}$ is the indicator function and $\tau$ a decision threshold.

\myparagraphdot{Training Updates for Malware Classifiers}
In realistic deployment settings, detectors are periodically updated as new data is collected to cope with evolving distributions.
We model malware detection over a temporally ordered sequence of data batches indexed by $t\in\{0,\dots,T\}$. At step $t$, $D^{(t)}$ denotes the labeled training set, and $U^{(t)}$ denotes the newly collected batch of samples whose labels are unknown at acquisition time. 
The detector trained at step $t$ is denoted as $f^{(t)}$ and outputs a malware score $f^{(t)}(\vct{x})\in[0,1]$.
For each update, AL selects a set to be labeled $D_{\mathrm{AL}}^{(t)}\subseteq U^{(t)}$, while SSL produces a pseudo-labeled set $D_{\mathrm{SSL}}^{(t)}\subseteq U^{(t)}$. The training set $D^{(t)}$ is then augmented with these newly-labeled samples, yielding $D^{(t+1)}$. 

\begin{figure*}[t]
    \centering
    \includegraphics[width=0.95\linewidth]{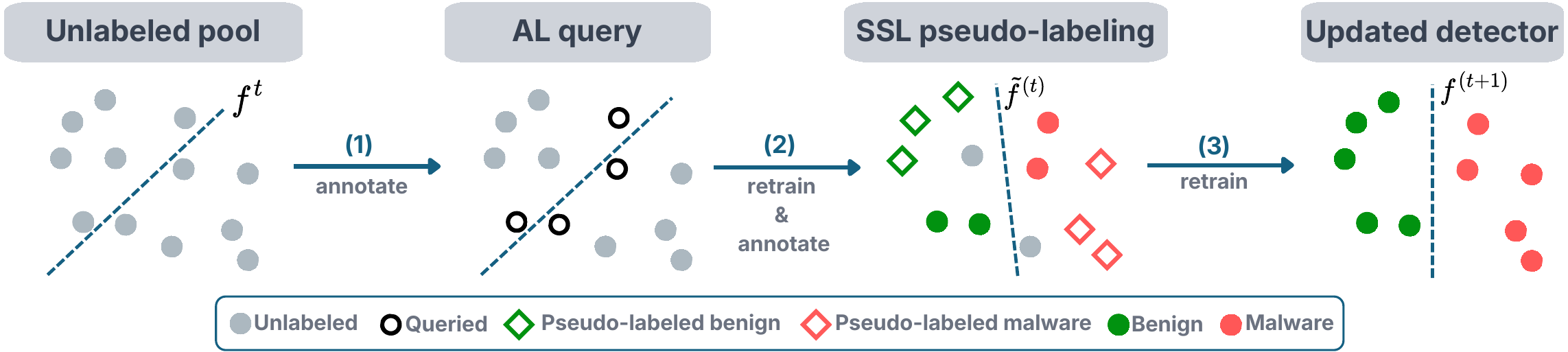}
    \caption{Overview of our proposed pipeline combining AL and SSL. The dash‑dot line represents the detector's decision boundary. Given a set of newly collected unlabeled samples (left), AL queries a small set of informative samples for expert annotation (center), while SSL automatically assigns pseudo-labels to high-confidence samples  (right), enabling efficient model retraining with reduced labeling cost.}
    \label{fig:al_ssl_pipeline}
\end{figure*}

\subsection{Active Learning Strategies}\label{Sect:AL}
AL selects a small number of unlabeled samples expected to provide the greatest benefit when updating the detector.
We focus on \emph{pool-based} approaches, where models access large pools of unlabeled samples and select the most informative ones for labeling, aligning with operational malware analysis, where suspicious software is continuously collected and selectively labeled offline under a constrained expert budget. We exclude \emph{stream-based} AL methods, where data comes sequentially and needs immediate labeling. Similarly, we do not consider \emph{query synthesis} AL techniques that request labels for both real and synthetically generated data, as synthetic samples may not correspond to valid samples in input space, potentially wasting oracle effort.
An AL strategy uses the prediction scores $f(\vct x)$ to rank the samples in $U^{(t)}$ according to its selection criterion and queries the top $k$ samples for oracle labeling, yielding $D_{\mathrm{AL}}^{(t)}$. 
We define a query function $q^{(t)}: \set X \to \{0, 1\}$ that selects the samples to be labeled:
\begin{equation}
D_{\mathrm{AL}}^{(t)} =
\left\{
(\vct x, y)
\;\middle|\;
\vct x \in U^{(t)},\;
q^{(t)}(\vct x) = 1
\right\},
\end{equation}
where $y$ refers to the label provided by the analyst.
The function $q$ is generally assigned through a method-specific score function, which is thresholded to obtain the selection:

\myparagraphdot{Random Sampling (RS)} 
In RS, instances are selected uniformly at random. Despite its simplicity, it serves as a baseline for evaluating the effectiveness of more advanced AL strategies.
Formally, given a query rate $\pi$, the query function for RS is defined using the Bernoulli distribution:
\begin{equation}
q_{\mathrm{RS}}^{(t)}(\vct x) \sim \mathrm{Bernoulli}(\pi)\,.
\end{equation}

\myparagraph{Margin Sampling (MS)~\cite{settles2009active}.}
MS selects the sample with the smallest difference between the top two predicted class probabilities.
In binary classification, the query function becomes:
\begin{equation}
q_{\mathrm{MS}}^{(t)}(\vct x) =
\mathbb{I}\!\left[
\bigl|2f^{(t)}(\vct x) - 1\bigr|
\le m_\mathrm{MS}
\right],\,
\end{equation}
where $m_\mathrm{MS}$ is the margin between the two class predictions.

\myparagraph{Least-Confident Sampling (LCS)~\cite{settles2009active}.}
LCS selects the instances with the lowest prediction confidence, such as:
\begin{equation}
q_{\mathrm{LCS}}^{(t)}(\vct x) =
\mathbb{I}\!\left[
\max\!\bigl(f^{(t)}(\vct x),\, 1 - f^{(t)}(\vct x)\bigr)
\le m_{\mathrm{LCS}}
\right]\,,
\end{equation}
where $m_\mathrm{LCS}$ is the margin for the minimum confidence. 
This function targets samples where the classifier is less confident in its decision, focusing on instances near the decision boundary. 

\myparagraph{Entropy Sampling (ES)~\cite{settles2009active}.}
ES selects the sample with the highest predictive entropy $\mathcal{H}$.
Its query function is defined as:
\begin{equation}
q_{\mathrm{ES}}^{(t)}(\vct x) =
\mathbb{I}\!\left[
\mathcal{H}\!\left(f^{(t)}(\vct x)\right)
\ge m_{\mathrm{ES}}
\right]\,,
\end{equation}
where $m_\mathrm{ES}$ is a threshold for the entropy. Entropy measures the total uncertainty of the model across all classes.
For binary linear classifiers, MS, LCS, and ES are equivalent, as they all select samples closest to the decision boundary $f(\vct x) = \tau$; consequently, they induce identical query rankings.

\myparagraph{Expected Average Precision (EAP)~\cite{wang2018uncertainty}.}
EAP estimates, for each sample, the expected improvement in Average Precision (AP) after retraining if the label is provided:
\begin{equation}
q_{\mathrm{EAP}}^{(t)}(\vct x) =
\mathbb{I}\!\left[
\sum_{y \in \{0,1\}}
f_y^{(t)}(\vct x) \,
\mathrm{AP}\!\left(f^{(t+1)} \mid (\vct x, y)\right)
\ge m_{\mathrm{EAP}}
\right]
\end{equation}
This strategy may become computationally costly due to the repeated retraining required to estimate improvement in AP.

\myparagraph{CLustering with Uncertainty-weighted Embeddings (CLUE)~\cite{prabhu2021activedomainadaptationclustering}.}
CLUE clusters samples in the model's embedding space and weights them based on their uncertainty, selecting the nearest samples to centroid clusters.
We apply it in feature space, defining the query function as:
\begin{equation}
q_{\mathrm{CLUE}}^{(t)}(\vct x) =
\mathbb{I}\!\left[
\mathcal{H}\!\left(f^{(t)}(\vct x)\right)
\cdot
\bigl\|\vct x - \boldsymbol{\mu}_{c(\vct x)}\bigr\|_2
\ge m_{\mathrm{CLUE}}
\right],
\end{equation}
where $\vct \mu_{c(\vct x)}$ is the centroid of the cluster to which $\vct x$ is assigned, the predictive entropy $\mathcal{H}(f^{(t)}(\vct x))$ defines each sample’s contribution to cluster formation, and $m_\mathrm{CLUE}$ is a threshold on the distance from the centroid.

\myparagraph{CoreSet Selection (CoreSet)~\cite{sener2017active}.}
CoreSet selects a representative subset that minimizes the distance between each sample and the selected set in embedding space (in our setting, the feature space). It constructs a batch $S_{\mathrm{CS}}^{(t)} \subset U^{(t)}$ by iteratively selecting samples that maximize their minimum distance to the already selected set, following a farthest-first criterion:
\begin{equation}
\vct x^* =
\arg\max_{\vct x \in U^{(t)} \setminus S}
\min_{\mathbf{z} \in S}
\left\lVert \vct x - \vct z \right\rVert_2,
\end{equation}
where $S$ is initialized either with the labeled set $D^{(t)}$ or with an empty set.
The process is repeated until the desired number of samples is selected.
The query function is defined as:
\begin{equation}
q_{\mathrm{CoreSet}}^{(t)}(\vct x) =
\mathbb{I}\!\left[\vct x \in S_{\mathrm{CS}}^{(t)}\right].
\end{equation}

\myparagraph{Batch Active Learning by Diverse Gradient Embeddings (BADGE)~\cite{ash2020badge}.}
BADGE is a batch active learning strategy that selects informative and diverse samples by operating in the gradient embedding space.
It computes the gradient embedding for each unlabeled sample $\vct x\in U^{(t)}$ as the gradients of the cross-entropy loss with respect to the model embeddings, using the label $\hat y(\vct x)$ predicted by the current detector $f^{(t)}$. We extend it to non-differentiable models by computing approximate gradient embeddings in the model feature space using a surrogate softmax linear classifier. For the $i$-th class, gradient embeddings can thus be computed as:
\begin{equation}
\psi_i(\vct x) = \bigl(f^{(t)}_i(\vct x)-\mathbb{I}(\hat y(\vct x)=i)\bigr) \vct x.
\end{equation}
BADGE then selects a batch $S_{\mathrm{BADGE}}^{(t)} \subset U^{(t)}$ by applying $k$-means++ seeding to the set of gradient embeddings.
The query function is then defined as:
\begin{equation}
q_{\mathrm{BADGE}}^{(t)}(\vct x) =
\mathbb{I}\!\left[\vct x \in S_{\mathrm{BADGE}}^{(t)}\right].
\end{equation}
\subsection{Semi-supervised Learning Strategies}\label{Sect:SSL}
SSL aims to reduce labeling cost by automatically assigning labels to a subset of newly collected samples before incorporating them into the training set. We adopt two SSL techniques relying on pseudo-labels to iteratively update the model, which are fully model-agnostic and require no modifications to the underlying detector, making them particularly suitable for our setting.
We build the SSL dataset through a pseudo-labeling function $g^{(t)}: \set X \to \{0, 1, \bot\}$, where $\bot$ means that no pseudo label is assigned and the sample is excluded from the set:
\begin{equation}
    D_{\mathrm{SSL}}^{(t)} =
    \left\{
    \bigl(\vct x, g^{(t)}(\vct x)\bigr)
    \;\middle|\;
    \vct x \in U^{(t)},\;
    g^{(t)}(\vct x) \neq \bot
    \right\}.
\end{equation}

\myparagraph{Self-Training with Symmetric Thresholding (ST)~\cite{yarowsky1995unsupervised}.}
ST builds the pseudo-labeled set by selecting high-confidence predictions from the unlabeled pool $U^{(t)}$ using the current detector score $f^{(t)}$. 
It defines the pseudo-labeling function as:
\begin{equation}
g_{\mathrm{ST}}^{(t)}(\vct x) =
\begin{cases}
1 & \text{if } f^{(t)}(\vct x) \ge \gamma, \\
0 & \text{if } f^{(t)}(\vct x) \le 1 - \gamma, \\
\bot & \text{otherwise},
\end{cases} 
\end{equation}
assigning a pseudo-label to each $\vct{x}\in U^{(t)}$ whose prediction confidence exceeds a fixed threshold $\gamma$, equal for both classes.
ST's main limitation is its inability to address class imbalance, common in malware detection, because a single threshold tends to favor the majority class (goodware), worsening the imbalance and reducing the detection of rare malware.

\myparagraph{Self-Training with Asymmetric Thresholding (AT)~\cite{stanescu2014asymmetric}.}
AT is a variant of ST that addresses class imbalance using class-specific confidence thresholds.
Its pseudo-labeling function is:
\begin{equation}
g_{\mathrm{AT}}^{(t)}(\vct x) =
\begin{cases}
1 & \text{if } f^{(t)}(\vct x) \ge \gamma^{+}, \\
0 & \text{if } f^{(t)}(\vct x) \le \gamma^{-}, \\
\bot & \text{otherwise},
\end{cases} 
\end{equation}
which pseudo-labels each sample using two different thresholds $\gamma^{-}$ and $\gamma^{+}$ for goodware and malware, respectively. By properly tuning them, AT can increase the fraction of pseudo-labeled malware and prevent $D_{\mathrm{SSL}}^{(t)}$ from being dominated by goodware, improving sensitivity to rare malicious behaviors.

\section{Label-efficient Malware Detector Updates}\label{Sect:methodology}

This section details our framework for label-efficient detector updates and introduces a feature-level drift analysis to measure the stability of discriminative features over time.

\subsection{Combining AL and SSL}\label{Sect:AL_SSL_combined}
We adopt a two-stage update procedure, shown in \autoref{fig:al_ssl_pipeline}, in which AL is applied before SSL, interleaved with a retraining step.
AL uses a limited labeling budget to select informative samples, while SSL subsequently enlarges the update set at negligible cost by pseudo-labeling high-confidence samples.
We apply AL before SSL so that pseudo-labeling relies on a partially adapted detector, reducing the risk of early errors being propagated and amplified over time.
At time step $t$, the detector $f^{(t)}$ is trained on the available labeled dataset $D^{(t)}$ and applied to a newly collected unlabeled batch $U^{(t)}$.
AL is first applied to $U^{(t)}$ to identify a subset of samples to label, yielding $D_{\mathrm{AL}}^{(t)}$.
We then update the training set with these newly labeled samples,
$\tilde D^{(t)} = D^{(t)} \cup D_{\mathrm{AL}}^{(t)}$,
and retrain the detector to obtain an intermediate model $\tilde f^{(t)}$.
SSL is then applied to the remaining unlabeled set,
$\tilde U^{(t)} = U^{(t)} \setminus  D_{\mathrm{AL}}^{(t)}$,
using the intermediate detector $\tilde f^{(t)}$ to assign pseudo-labels to high-confidence instances, yielding a pseudo-labeled set $D_{\mathrm{SSL}}^{(t)}$.
By construction, AL and SSL operate on disjoint subsets of samples, with AL allocating the expensive labeling budget to informative samples and SSL supporting the update with low-cost pseudo-labels.
Finally, the updated model $f^{(t+1)}$ is obtained by training $f$ on the enriched training set, defined as:
\begin{equation}
D^{(t+1)} = D^{(t)} \cup D_{\mathrm{AL}}^{(t)} \cup D_{\mathrm{SSL}}^{(t)}.
\end{equation}
This update procedure is then repeated over successive time steps, enabling the detector to adapt to evolving malware distributions while limiting manual labeling effort.

\myparagraphdot{Practical Implications} 
This design jointly achieves robustness to distribution drift, label efficiency, and model-agnosticity.
Our pipeline is compatible with both full-history and sliding-window updates.
In long-running deployments, $D^{(t)}$ may be maintained as a sliding window, discarding the oldest samples and retaining only the most recent ones. 
This keeps retraining scalable and avoids older retained samples from accumulating and numerically dominating the training set.
As the data volume in our experiments is manageable, we retain the full history when updating the detector (i.e., no samples are discarded).
Our approach is label-efficient: low AL labeling budgets reduce the need for time-consuming analysis, whereas SSL pseudo-labels can be generated at scale at negligible cost.
We also recall that the pipeline is model-agnostic: the AL and SSL components use only the detector outputs (and, for some AL strategies, feature-space representations) without modifying their architecture or training procedure.

\subsection{Drift Analysis}\label{Sect:drift}
We analyze the distribution drift based on the temporal stability of detector-agnostic feature--class associations.

\myparagraphdot{Motivation}
Unlike analyses based on model-specific latent spaces or architecture-dependent signals, we operate on feature values and class labels, enabling comparisons across different detectors with the same feature representation.
Our goal is to quantify whether features that are discriminative on the current training set, \ie significantly associated with one class (goodware or malware), remain stable over time, \ie preserve the same class association in subsequent batches (e.g., in the next month for EMBER or in the next quarter for ELSA).
Intuitively, if a detector is trained using feature--class associations that no longer hold over time, then its predictions rely on stale cues, and performance is expected to degrade.

\myparagraphdot{Methodology}
To quantify feature--class association, we analyze each feature $j$ separately, comparing their values in both classes by applying the Wilcoxon--Mann--Whitney (WMW) test and computing its normalized $U$ statistic, equivalently the ROC AUC, defined as $\mathrm{AUC}_j$. We use WMW as it provides (i)~a measure of statistical significance through the $p$-value $p_j$, assessing whether the observed association is due to random variation, and (ii)~an association direction, indicating whether a feature can be associated with a class.
Based on that, we define for each feature the \emph{class-association indicator}, encoding the direction of the association only when statistically significant:
\begin{equation}
a_j \;=\;
\begin{cases}
\mathrm{sgn}\!\left(2\mathrm{AUC}_j - 1\right), & \text{if } p_j < 0.05,\\[4pt]
0, & \text{otherwise}.
\end{cases}
\end{equation}
Thus, for statistically significant associations, $a_j$ encodes their direction, i.e., $+1$ for malware and $-1$ for goodware, respectively. Otherwise, $a_j=0$ indicates that no reliable association is found, either because it is not statistically significant or no clear direction emerges ($\mathrm{AUC}_j \approx 0.5$).
At each time step $t$, we compute the indicator on train and test data under the same feature representation, obtaining $a^{(t)}_{tr,j}$ and $a^{(t)}_{ts,j}$ for each feature $j$, respectively.
Discrepancy between train and test association directions reveals the presence of distribution drift, while preservation shows its absence. On the other hand, no assumption can be made when the association cannot be determined in both the train and test features.
Based on those assumptions, we compute the \emph{stability score} metric as:
\begin{equation}
\beta^{(t)}=\frac{1}{d}\sum_{j=1}^{d}\Big(a^{(t)}_{tr,j} \cdot a^{(t)}_{ts,j}- \mathbb{I}\!\Big[\big|a^{(t)}_{tr,j}\big|+\big|a^{(t)}_{ts,j}\big|=1\Big]\Big).
\end{equation}
Features whose class association is statistically significant and is preserved between train and test data contribute positively to $\beta^{(t)}$ through the first summation term, as their presence is desirable. Conversely, features that change association direction contribute negatively, through the first (if the associations have opposite directions) and the second (if one of the two indicators is zero) summation terms.
Consequently, larger $\beta^{(t)}$ indicates greater stability, whereas lower values indicate misalignment between the detectors' feature representations and the discriminative features characterizing the current data distribution.
This also provides a detector-agnostic way to interpret AL/SSL updates: successful sample selection should increase the fraction of stable discriminative features, thereby mitigating performance decay under distribution drift.

\section{Experiments}
\label{Sect:experiments}
We first describe the experimental setup and evaluation protocol, then present and discuss the results.

\subsection{Experimental Setup}
\myparagraphdot{Datasets} 
We consider the EMBER Windows dataset~\cite{anderson2018ember} and the Android ELSA dataset~\cite{ramd}.
EMBER provides $2,381$ features extracted from PE files. We combine the 2017 and 2018 releases and exclude unlabeled samples, obtaining $1,500,000$ PE files that are approximately balanced overall. However, under the temporal split, the class prior varies across months, inducing batch-level imbalance. Moreover, concatenating two releases introduces an additional distribution change at the year boundary, due to different sampling strategies. To emulate a realistic deployment setting, we adopt a monthly stream: samples from the first month are used for training only, while samples from subsequent months form consecutive incoming unlabeled batches $U^{(t)}$. Each $U^{(t)}$ is first used to evaluate $f^{(t)}$ and then processed by the AL--SSL pipeline to obtain $D^{(t+1)}$.
The ELSA dataset comprises applications collected from AndroZoo~\cite{Allix:2016:ACM:2901739.2903508} with pre-extracted Drebin features~\cite{arp2014drebin}. The dataset is strongly imbalanced, with a 9:1 benign-to-malware ratio. We use data from 2017 to 2021 for training (75,000 applications). The remaining 137,500 applications, sampled between Jan.~2020 and Jun.~2022, are reserved for evaluation and split into quarterly batches. As in~\cite{chen2023continuous}, we retain the $10,000$ most frequent features to reduce computational overhead; despite the fixed representation, the set of \emph{active} (non-zero) features and their discriminative patterns vary over time, reflecting the evolution of app behaviors.

\myparagraphdot{Models}
On ELSA, we rely on a linear Support Vector Machine (SVM) with $C = 0.1$, and a Random Forest (RF) with $80$ trees and maximum depth $30$. On EMBER, we use only RF, as training an SVM is not feasible in several of the considered settings due to memory constraints. 
As AL strategies rely on probability estimates, we apply Platt scaling to the SVM, thereby facilitating uncertainty-based active learning methods.

\myparagraphdot{Evaluation Protocol} To assess the effectiveness of the considered retraining strategies, we use the F1 score and the Recall score when fixing the False Positive Rate (FPR) to $1\%$.
We compare them against two baselines: a \textit{no-retraining} (NR) model trained only once on the initial set $D^{(0)}$ and never updated; and a \textit{full-labeling} (FL) model retrained at each step $t$ using the fully labeled dataset $D^{(t)} \cup \left\{
    \bigl(\vct x, y\bigr)
    \;\middle|\;
    \vct x \in U^{(t)}
    \right\}$.
All the tested AL and SSL methods are trained with full labeling only on $D^{(0)}$ and are retrained afterwards using the dataset (with labels) obtained via the selected strategies at time $t$.

\myparagraphdot{Label budgets}
We control the labeling budget as a percentage of the unlabeled set. For each method, thresholds and selection criteria are tuned to meet the corresponding budget, ensuring that all approaches label an equal number of samples per step.

\subsection{Experimental Results on AL, SSL, and AL+SSL}
\label{Sect:results_SSL_AL}

\begin{figure*}[t]

    
    \makebox[\textwidth][c]{%
        \hspace{-4em}
        \includegraphics[width=.98\textwidth]{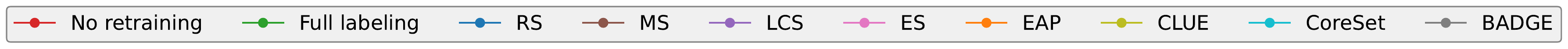}%
        \hspace{-4em}
    }
    \makebox[\textwidth][c]{%
        \hspace{-4em}
        \subfigure[ELSA – AL Comparison (Budget - 10\%)]{
            \includegraphics[trim=0cm 0cm 0cm 0cm, clip, width=0.48\textwidth]{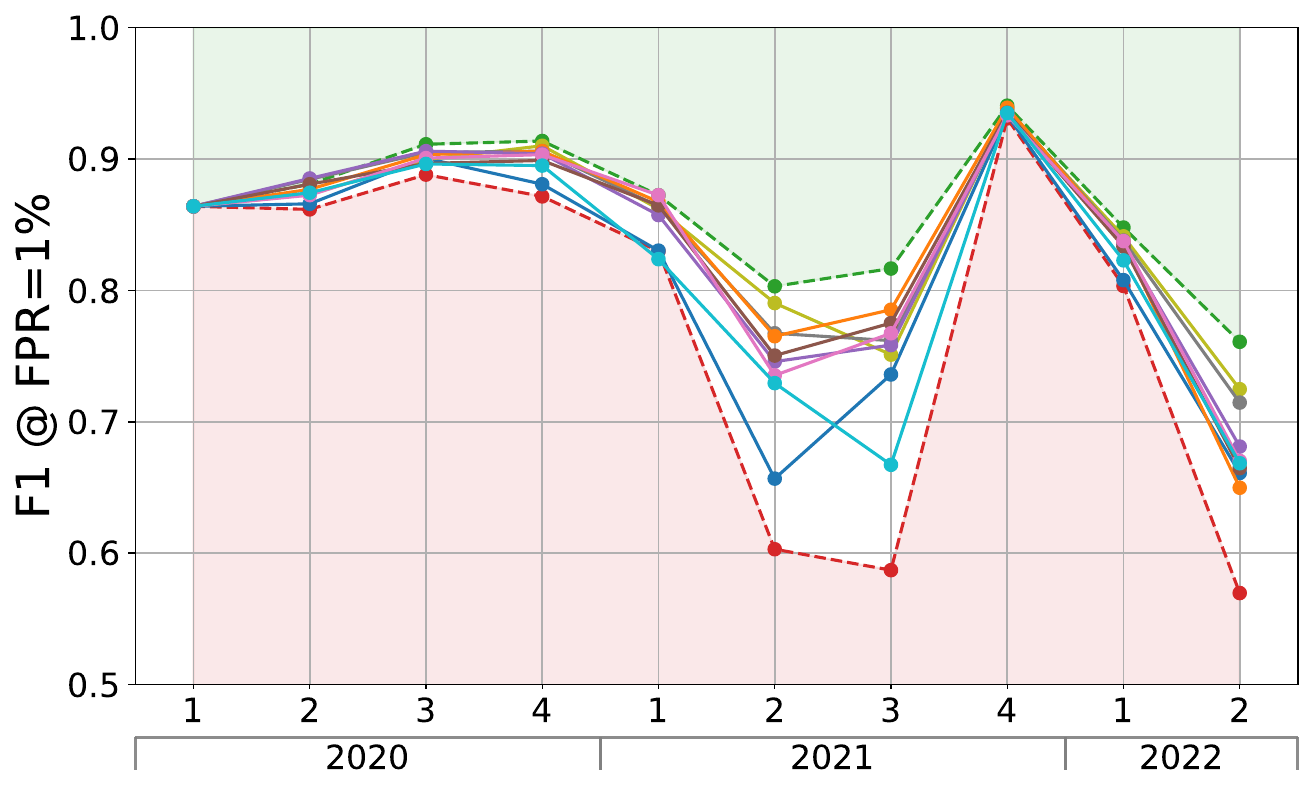}
            \label{fig:ELSA_AL_f1_budget_10}
        }
        \hspace{-1em}
        \subfigure[EMBER – AL Comparison (Budget - 10\%)]{
            \includegraphics[trim=1cm 0cm 0cm 0cm, clip, width=0.46\textwidth]{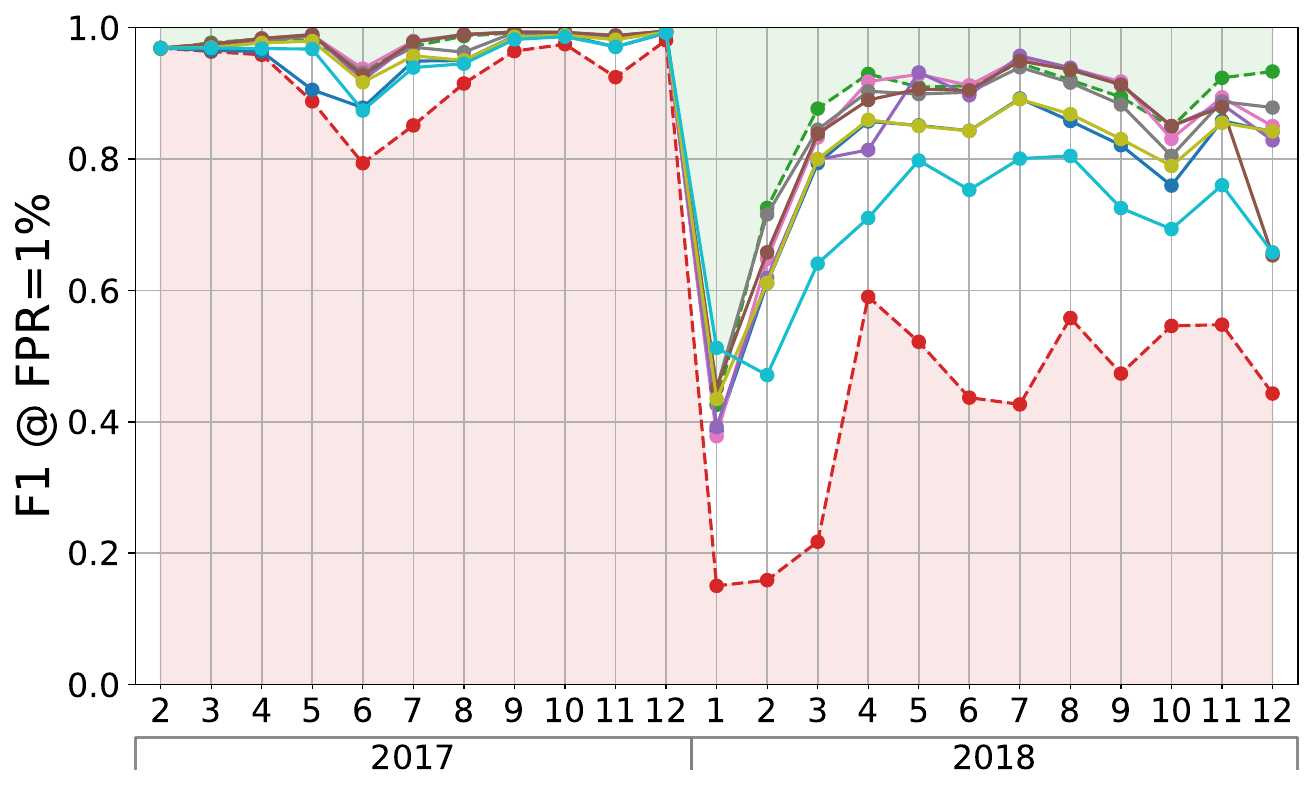}
            \label{fig:EMBER_AL_f1_budget_10}
        }
        \hspace{-4em}
    }

    \makebox[\textwidth][c]{%
        \hspace{-4em}
        \includegraphics[width=0.48\textwidth]{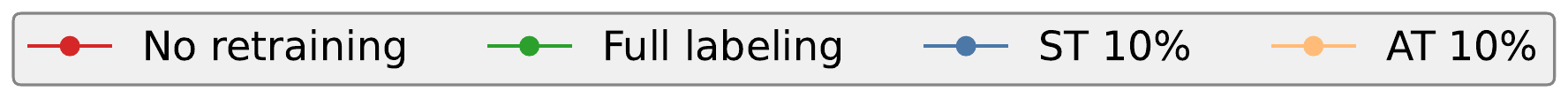}%
        \hspace{-4em}
    }
    \makebox[\textwidth][c]{%
        \hspace{-4em}
        \subfigure[ELSA – SSL Comparison (Budget - 10\%)]{
            \includegraphics[trim=0cm 0cm 0cm 0cm, clip, width=0.48\textwidth]{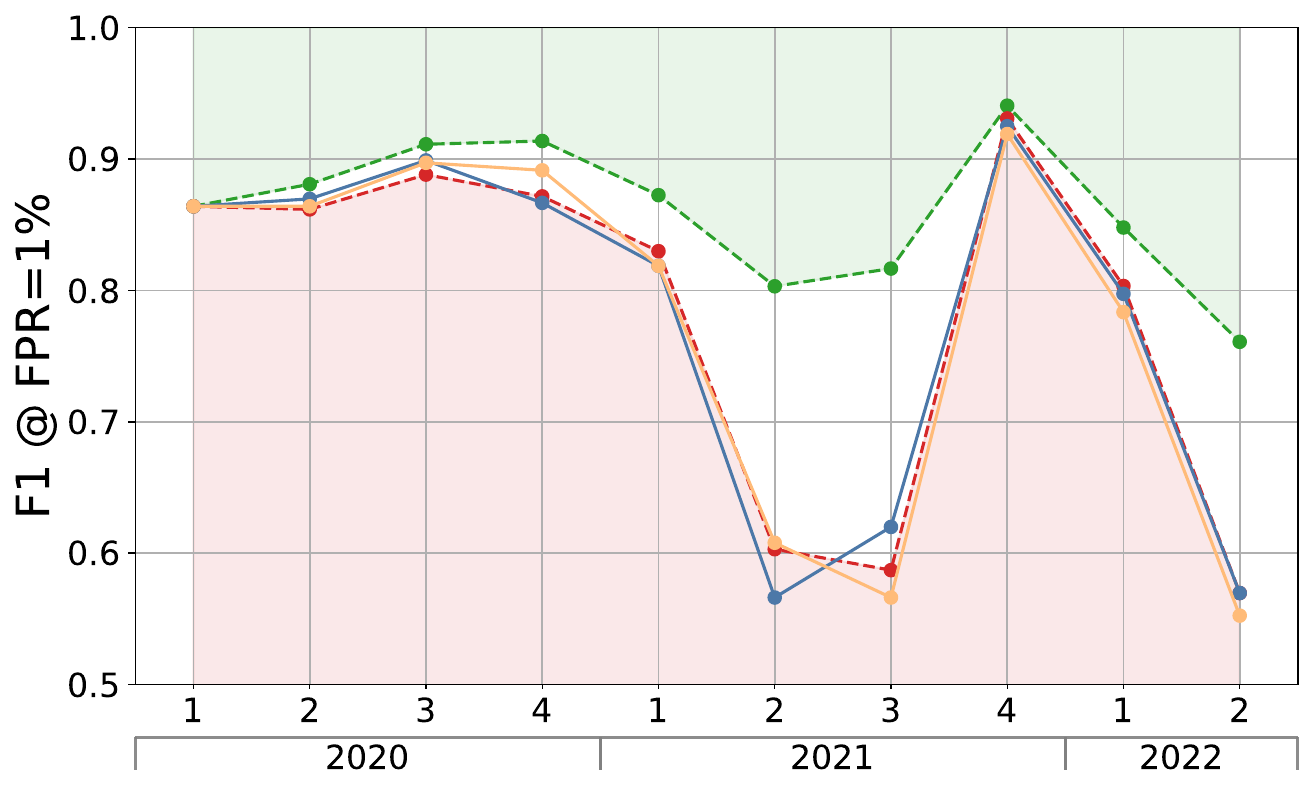}
            \label{fig:ELSA_SSL_f1_budget_10}
        }
        \hspace{-1em}
        \subfigure[EMBER – SSL Comparison (Budget - 10\%)]{
            \includegraphics[trim=1cm 0cm 0cm 0cm, clip, width=0.46\textwidth]{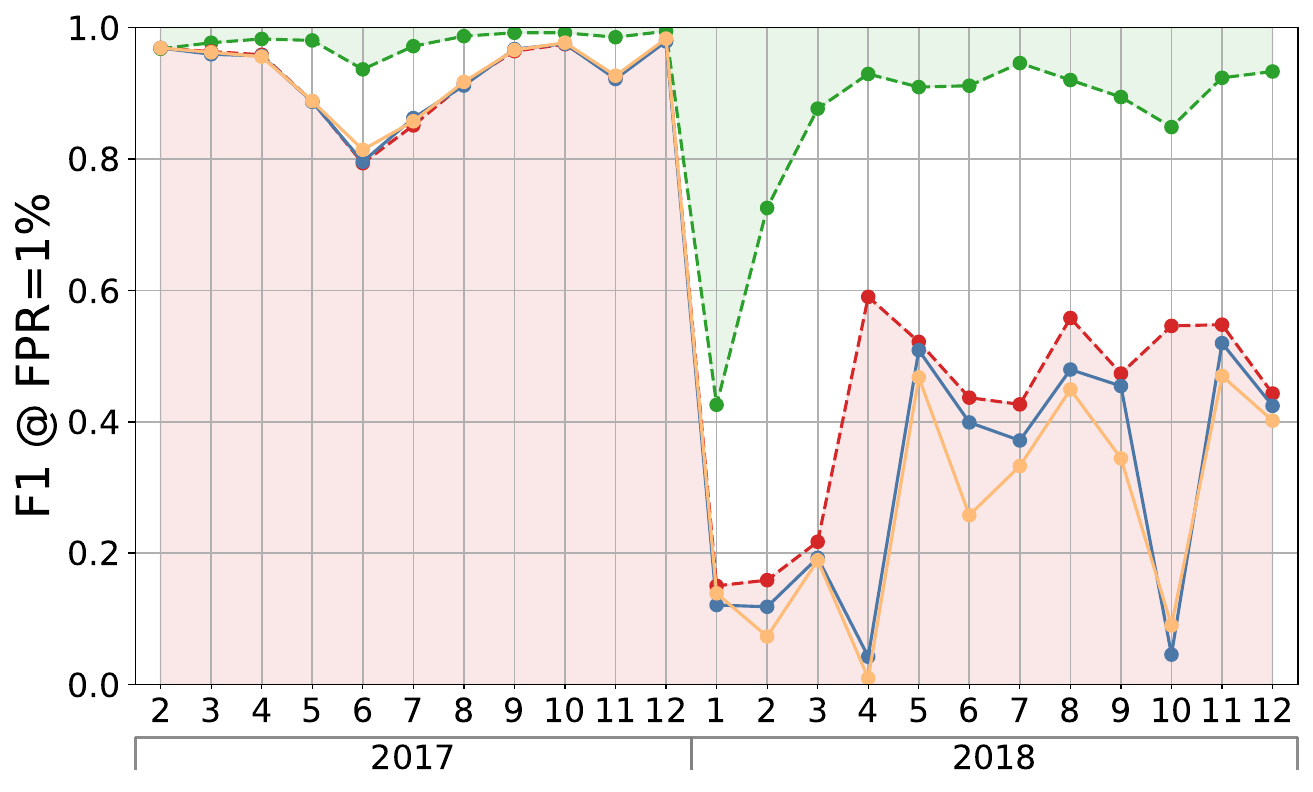}
            \label{fig:EMBER_SSL_f1_budget_10}
        }
        \hspace{-4em}
    }
    \caption{F1 score at FPR \(=1\%\) over time for RF on ELSA (quarterly) and EMBER (monthly) with a \(10\%\) labeling budget. Top: comparison of AL strategies, alongside NR and FL references. Bottom: comparison of SSL strategies under the same budget, with the same references.}
    \label{fig:AL_SSL_f1_budgets}
\end{figure*}

\myparagraphdot{AL Analysis}
We first evaluate AL in isolation to assess whether selective labeling can support detector updates under temporal drift with limited annotation budgets. We compare the strategies described in~\autoref{Sect:AL} across four label budgets, approximately $1\%$, $2\%$, $5\%$, and $10\%$ of the incoming batch. Higher budgets are not considered, as our goal is to focus on the low-label regime, where annotations are constrained, and the benefit of selective querying is most informative. We exclude EAP from EMBER evaluation as it requires retraining per sample and cannot scale to large datasets due to time and memory constraints.
Figs.~\ref{fig:ELSA_AL_f1_budget_10}~and~\ref{fig:EMBER_AL_f1_budget_10} show the F1-score evolution for the AL techniques in both datasets with a fixed labeling budget of $10\%$ of the total samples. The results clearly indicate that detectors can be effectively retrained without labeling all available samples, reporting consistent performance across both datasets. Specifically, when fixing a budget of $10\%$, most techniques exhibit a trend that closely matches the performance of FL, which uses the entire dataset for retraining. This demonstrates that AL techniques can reduce the amount of labels required by $90\%$, helping to counteract the performance decline caused by distribution drift in both ELSA and EMBER datasets, and are generally effective and not limited to specific models, in contrast to prior claims~\cite{chen2023continuous}.
Tables~\ref{tab:AL_avg_metrics}~and~\ref{tab:EMBER_AL_avg_metrics} report detailed metrics across different budgets, showing that the data selection method in AL significantly affects performance, particularly with smaller budgets.
Both datasets report the same trends. BADGE is the overall best-performing method, especially in low-budget regimes, where more sophisticated strategies outperform simpler ones such as LSC or ES. 
As the budget increases, the gap between different strategies narrows, and even RS progressively approaches FL, since most methods have already queried enough samples to achieve high performance. 
At very low budgets, retraining can be high-variance: with only a few queried points, some AL strategies may select an unrepresentative batch (often dominated by the majority class or containing outliers). Incorporating these labels can perturb the decision function, resulting in lower performance than the NR baseline.
\begin{insightbox}
    AL can significantly reduce labeling costs while still achieving performance comparable to full labeling.
\end{insightbox}
\begin{table}[t]
  \centering
  \caption{Average recall ($\rho$) and F1 score across different AL strategies and label budgets, at fixed FPR of 1\%, for SVM and RF models on ELSA. Best overall results are in \textbf{bold}, the top AL method is \underline{underlined}, and values below the NR baseline are in \textcolor{red}{red}.}
  \resizebox{\columnwidth}{!}{%
  \begin{tabular}{l|cc|cc|cc|cc}
    \toprule
    \textbf{AL Strategy} & \multicolumn{2}{c|}{\textbf{1\% Budget}} & \multicolumn{2}{c|}{\textbf{2\% Budget}} & \multicolumn{2}{c|}{\textbf{5\% Budget}} & \multicolumn{2}{c}{\textbf{10\% Budget}} \\
    \midrule
    & $\rho$ & F1 & $\rho$ & F1 & $\rho$ & F1 & $\rho$ & F1 \\
    \midrule
    SVM + NR & 74.9 & 80.0 & 74.9 & 80.0 & 74.9 & 80.0 & 74.9 & 80.0 \\
    SVM + FL & \textbf{82.2} & \textbf{85.4} & \textbf{82.2} & \textbf{85.4} & \textbf{82.2} & \textbf{85.4} & 82.2 & 85.4 \\
    \midrule
    SVM + RS & \textcolor{red}{72.9} & \textcolor{red}{78.5} & \textcolor{red}{74.5} & \textcolor{red}{79.9} & 76.0 & 81.2 & 76.6 & 81.6 \\
    SVM + BADGE & \underline{78.6} & \underline{82.9} & 80.2 & 84.0 & 81.5 & 85.0 & \textbf{\underline{83.5}} & \textbf{\underline{86.3}} \\
    SVM + CLUE & 78.1 & 82.6 & \underline{80.4} & \underline{84.2} & \underline{82.2} & \underline{85.4} & 82.5 & 85.6 \\
    SVM + CoreSet & \textcolor{red}{74.8} & \textcolor{red}{79.9} & 76.1 & 81.0 & 75.6 & 80.6 & 79.0 & 83.2 \\
    SVM + EAP & 77.9 & 82.4 & 80.0 & 84.0 & 81.7 & 85.1 & 82.9 & 86.0 \\
    SVM + ES & 76.0 & 80.9 & 78.6 & 82.9 & 80.6 & 84.4 & 82.3 & 85.5 \\
    SVM + LCS & 76.0 & 80.9 & 78.6 & 82.9 & 80.6 & 84.4 & 82.3 & 85.5 \\
    SVM + MS & 76.0 & 80.9 & 78.6 & 82.9 & 80.6 & 84.4 & 82.3 & 85.5 \\
    \toprule
    RF + NR & 71.9 & 78.1 & 71.9 & 78.1 & 71.9 & 78.1 & 71.9 & 78.1 \\
    RF + FL & \textbf{83.1} & \textbf{86.1} & \textbf{83.1} & \textbf{86.1} & \textbf{83.1} & \textbf{86.1} & \textbf{83.1} & \textbf{86.1} \\
    \midrule
    RF + RS & \textcolor{red}{71.6} & \textcolor{red}{78.0} & 74.3 & 80.1 & 75.1 & 80.7 & 76.0 & 81.4 \\
    RF + BADGE & 74.4 & 80.1 & \underline{78.6} & \underline{83.1} & \underline{79.8} & \underline{84.0} & 80.6 & 84.4 \\
    RF + CLUE & \underline{75.2} & \underline{80.8} & 77.0 & 82.1 & 78.4 & 83.0 & \underline{80.6} & \underline{84.5} \\
    RF + CoreSet & \textcolor{red}{69.1} & \textcolor{red}{76.0} & \textcolor{red}{68.9} & \textcolor{red}{75.6} & 73.7 & 79.5 & 76.7 & 81.8 \\
    RF + EAP & 73.3 & 79.2 & 75.9 & 81.2 & 78.6 & 83.1 & 79.8 & 83.9 \\
    RF + ES & 72.5 & 78.4 & 72.4 & 78.5 & 77.1 & 82.1 & 79.3 & 83.6 \\
    RF + LCS & 72.3 & 78.5 & 74.3 & 80.0 & 76.4 & 81.5 & 79.5 & 83.7 \\
    RF + MS & \textcolor{red}{69.5} & \textcolor{red}{76.3} & 72.9 & 78.9 & 77.2 & 82.2 & 79.4 & 83.6 \\
    \bottomrule
  \end{tabular}}
  \label{tab:AL_avg_metrics}
\end{table}
\begin{table}[t]
  \centering
    \caption{Comparison of AL strategies for different label budgets on EMBER (RF). Refer to \autoref{tab:AL_avg_metrics} caption for metrics and formatting.}
  \resizebox{\columnwidth}{!}{%
  \begin{tabular}{l|cc|cc|cc|cc}
        \toprule
    \textbf{AL Strategy} & \multicolumn{2}{c|}{\textbf{1\% Budget}} & \multicolumn{2}{c|}{\textbf{2\% Budget}} & \multicolumn{2}{c|}{\textbf{5\% Budget}} & \multicolumn{2}{c}{\textbf{10\% Budget}} \\
    \midrule
    & $\rho$ & F1 & $\rho$ & F1 & $\rho$ & F1 & $\rho$ & F1 \\
    \midrule
    RF + NR & 56.6 & 66.3 & 56.6 & 66.3 & 56.6 & 66.3 & 56.6 & 66.3 \\
    RF + FL & \textbf{86.7} & \textbf{91.4} & \textbf{86.7} & \textbf{91.4} & \textbf{86.7} & \textbf{91.4} & \textbf{86.7} & \textbf{91.4} \\
    \midrule
    RF + RS & 67.0 & 76.7 & 72.4 & 81.4 & 75.1 & 83.1 & 79.0 & 86.5 \\
    RF + Badge & \underline{69.0} & \underline{78.3} & \underline{73.5} & \underline{81.9} & \underline{79.6} & \underline{86.5} & 84.8 & 90.3 \\
    RF + CLUE & 61.7 & 71.3 & 67.6 & 76.8 & 74.9 & 83.1 & 80.7 & 87.6 \\
    RF + CoreSet & \textcolor{red}{55.7} & \textcolor{red}{64.8} & \textcolor{red}{56.0} & \textcolor{red}{64.8} & 64.7 & 74.1 & 73.2 & 82.1 \\
    RF + ES & 59.4 & 68.3 & 63.0 & 71.4 & 78.2 & 85.3 & \underline{85.5} & \underline{90.4} \\
    RF + LCS & 59.8 & 69.0 & 63.3 & 71.4 & 78.1 & 84.9 & 84.2 & 89.6 \\
    RF + MS & 58.9 & 67.6 & 63.6 & 72.2 & 77.1 & 83.8 & 84.2 & 89.6 \\
    \bottomrule
  \end{tabular}}
  \label{tab:EMBER_AL_avg_metrics}
\end{table}

\myparagraphdot{SSL Analysis}
We next evaluate SSL in isolation to assess whether pseudo-labeling alone can support detector updates under temporal drift. We compare the strategies described in~\autoref{Sect:SSL} across pseudo-label budgets of $10\%$, $20\%$, $60\%$, and $80\%$. 
Higher budgets are excluded to prevent error accumulation from misclassified pseudo-labels.
In AT, we meet desired budgets by first pseudo-labeling the top $80\%$ most confident samples as malware and filling the remaining with the most confident goodware to obtain $(\gamma^+,\gamma^-)$.
Figs.~\ref{fig:ELSA_SSL_f1_budget_10}~and~\ref{fig:EMBER_SSL_f1_budget_10} report the F1-score evolution for each method in both datasets, under a pseudo-labeling budget of $10\%$. We observe that SSL techniques generally do not improve performance and even perform worse than the NR baseline, and are far from matching FL performance. 
Further insights can be drawn from Tables~\ref{tab:SSL_avg_metrics}~and~\ref{tab:EMBER_SSL_avg_metrics}, which summarize average performance metrics across different budgets. The results confirm the limited effectiveness of these strategies in both datasets. Furthermore, we observe a consistent trend: increasing the budget often raises the risk of misclassification, decreasing overall effectiveness.
On ELSA, with SVM, AT uses class-specific thresholds, as described in \autoref{Sect:SSL}, which can be overly permissive when malware scores overlap with benign scores, injecting noisy positive pseudo-labels. This is particularly harmful at (FPR = 1\%) and can shift the linear decision boundary farther than ST. RF is more robust, so AT can help relative to ST, although SSL remains below the baselines on both datasets. On EMBER, AT and ST perform similarly with RF, but still underperform the baseline.
Overall, these results demonstrate the limitations of relying solely on SSL: the quality of pseudo-labels is constrained by the reliability of models' predictions on new samples, which, in turn, deteriorates when these pseudo-labels are fed back into the retraining process. This reinforces existing biases and propagates errors, undermining the models' performance.
\begin{insightbox}
Under distribution drift, using SSL alone yields worse results than not retraining at all.
\end{insightbox}
\begin{table}[t]
  \centering
    \caption{Average recall ($\rho$) and F1 score across different SSL strategies and pseudo-label budgets, at fixed FPR of 1\%, for SVM and RF models on ELSA. Best overall results are in \textbf{bold}, the top SSL method is \underline{underlined}, and values below the NR baseline are in \textcolor{red}{red}.}

  \resizebox{1\columnwidth}{!}{%
  \begin{tabular}{l|cc|cc|cc|cc}
    \toprule
    \textbf{SSL strategy} & \multicolumn{2}{c|}{\textbf{10\% Budget}} & \multicolumn{2}{c|}{\textbf{20\% Budget}} & \multicolumn{2}{c|}{\textbf{60\% Budget}} & \multicolumn{2}{c}{\textbf{80\% Budget}} \\
    \midrule
    & $\rho$ & F1 & $\rho$ & F1 & $\rho$ & F1 & $\rho$ & F1 \\
    \midrule
    SVM + NR & 74.9 & 80.0 & 74.9 & 80.0 & 74.9 & 80.0 & 74.9 & 80.0 \\
    SVM + FL & \textbf{82.2} & \textbf{85.4} & \textbf{82.2} & \textbf{85.4} & \textbf{82.2} & \textbf{85.4} & \textbf{82.2} & \textbf{85.4} \\
    \midrule
    SVM + ST & \underline{\textcolor{red}{73.8}} & \underline{\textcolor{red}{79.1}} & \underline{\textcolor{red}{73.7}} & \underline{\textcolor{red}{79.0}} & \underline{\textcolor{red}{73.5}} & \underline{\textcolor{red}{78.9}} & \underline{\textcolor{red}{73.9}} & \underline{\textcolor{red}{79.3}} \\
    SVM + AT & \textcolor{red}{72.1} & \textcolor{red}{77.9} & \textcolor{red}{71.0} & \textcolor{red}{76.9} & \textcolor{red}{66.0} & \textcolor{red}{71.7} & \textcolor{red}{65.3} & \textcolor{red}{70.7} \\
    \midrule
    RF + NR & 71.9 & 78.1 & 71.9 & 78.1 & 71.9 & 78.1 & 71.9 & 78.1 \\
    RF + FL & \textbf{83.1} & \textbf{86.1} & \textbf{83.1} & \textbf{86.1} & \textbf{83.1} & \textbf{86.1} & \textbf{83.1} & \textbf{86.1} \\
    \midrule
    RF + ST & \underline{\textcolor{red}{71.8}} & \underline{\textcolor{red}{78.0}} & \textcolor{red}{69.0} & \textcolor{red}{75.8} & \textcolor{red}{70.9} & \textcolor{red}{77.1} & \textcolor{red}{67.9} & \textcolor{red}{75.0} \\
    RF + AT & \textcolor{red}{71.4} & \textcolor{red}{77.6} & \underline{\textcolor{red}{71.4}} & \underline{\textcolor{red}{77.6}} & \underline{\textcolor{red}{71.7}} & \underline{\textcolor{red}{77.9}} & \underline{\textcolor{red}{70.8}} & \underline{\textcolor{red}{77.3}} \\
    \bottomrule
  \end{tabular}}
  \label{tab:SSL_avg_metrics}
\end{table}
\begin{table}[t]
  \centering
    \caption{Comparison of SSL strategies for different pseudo-label budgets on EMBER (RF). Refer to \autoref{tab:SSL_avg_metrics} caption for metrics and formatting.}
  \resizebox{\columnwidth}{!}{%
  \begin{tabular}{l|cc|cc|cc|cc}
        \toprule
    \textbf{SSL strategy} & \multicolumn{2}{c|}{\textbf{10\% Budget}} & \multicolumn{2}{c|}{\textbf{20\% Budget}} & \multicolumn{2}{c|}{\textbf{60\% Budget}} & \multicolumn{2}{c}{\textbf{80\% Budget}} \\
    \midrule
    & $\rho$ & F1 & $\rho$ & F1 & $\rho$ & F1 & $\rho$ & F1 \\
    \midrule
    RF + NR & 56.6 & 66.3 & 56.6 & 66.3 & 56.6 & 66.3 & 56.6 & 66.3 \\
    RF + FL & \textbf{86.7} & \textbf{91.4} & \textbf{86.7} & \textbf{91.4} & \textbf{86.7} & \textbf{91.4} & \textbf{86.7} & \textbf{91.4} \\
    \midrule
    RF + ST & \underline{\textcolor{red}{52.1}} & \underline{\textcolor{red}{60.3}} & \underline{\textcolor{red}{52.3}} & \underline{\textcolor{red}{60.9}} & \underline{\textcolor{red}{48.1}} & \underline{\textcolor{red}{55.7}} & \textcolor{red}{48.1} & \underline{\textcolor{red}{56.1}} \\
    RF + AT & \textcolor{red}{50.8} & \textcolor{red}{58.5} & \textcolor{red}{51.6} & \textcolor{red}{59.8} & \textcolor{red}{47.4} & \textcolor{red}{54.6} & \underline{\textcolor{red}{48.3}} & \textcolor{red}{56.0} \\
    \bottomrule
  \end{tabular}}
  \label{tab:EMBER_SSL_avg_metrics}
\end{table}
\begin{figure*}[t]
    \makebox[\textwidth][c]{%
        \hspace{-4em}
        \subfigure[ELSA – BADGE (Budget - 1\%)]{
            \includegraphics[trim=0cm 0cm 0cm 0cm, clip, width=0.48\textwidth]{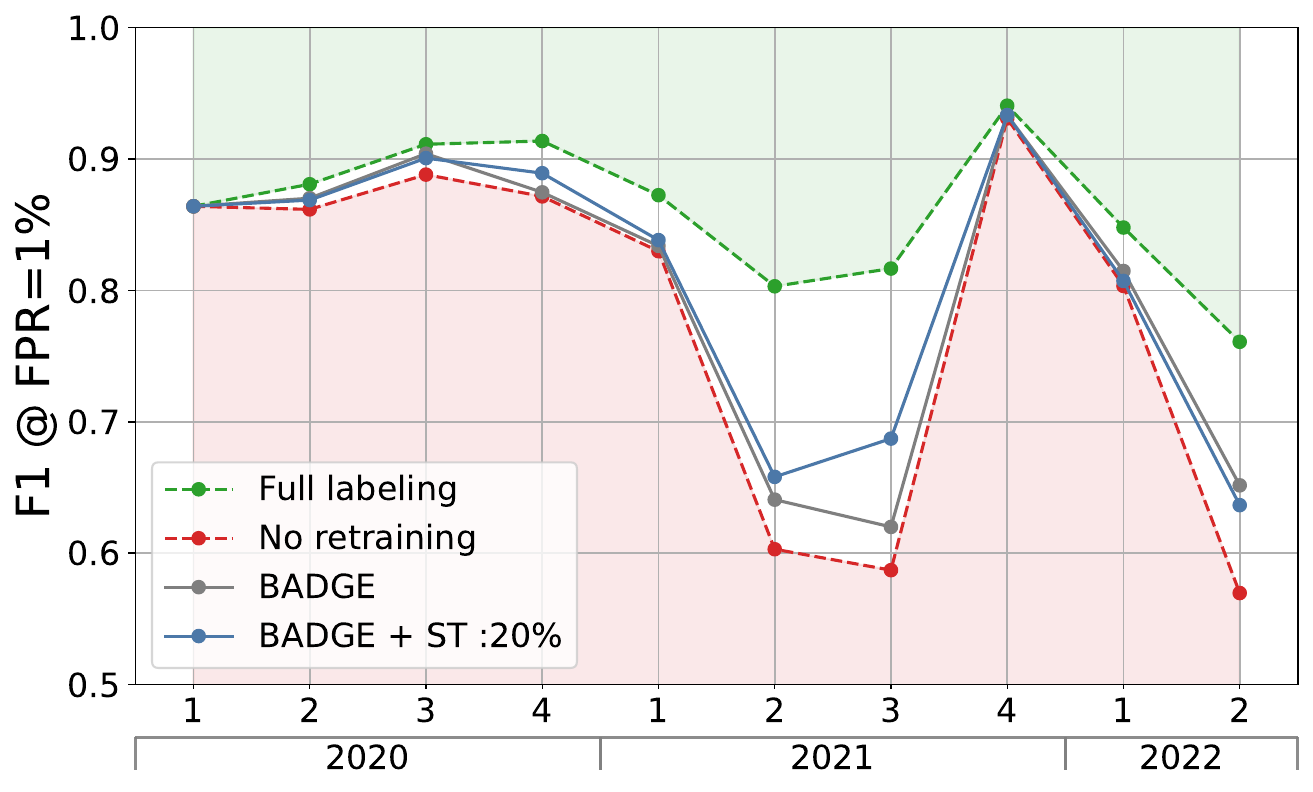}
            \label{fig:AL_SSL_ELSA_BADGE}
        }
        \hspace{-1em}
        \subfigure[EMBER – BADGE (Budget - 2\%)]{
            \includegraphics[trim=1cm 0cm 0cm 0cm, clip, width=0.46\textwidth]{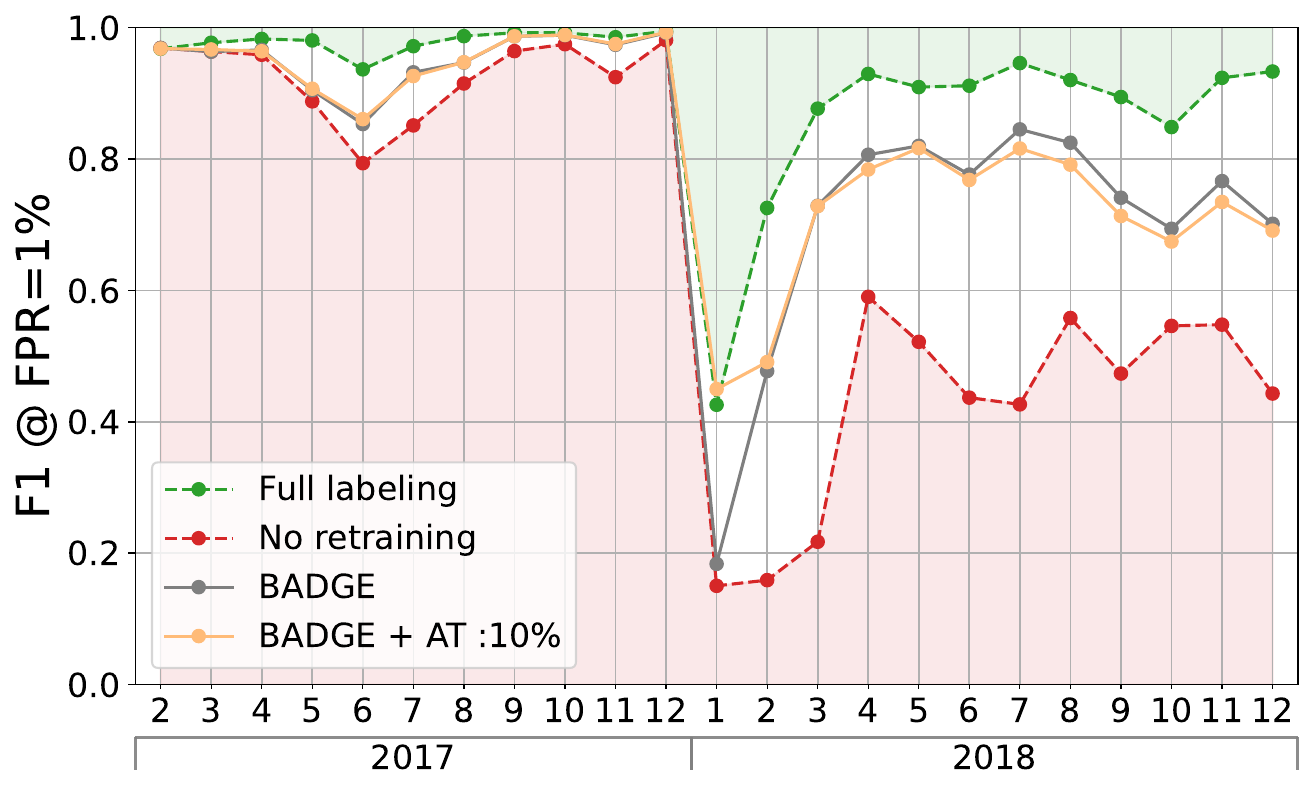}
            \label{fig:AL_SSL_EMBER_BADGE}
        }
        \hspace{-4em}
    }
    \\
    \makebox[\textwidth][c]{%
        \hspace{-4em}
                \subfigure[ELSA – BADGE (Budget - 1\%)]{
            \includegraphics[trim=0cm 0cm 0cm 0cm, clip, width=0.48\textwidth]{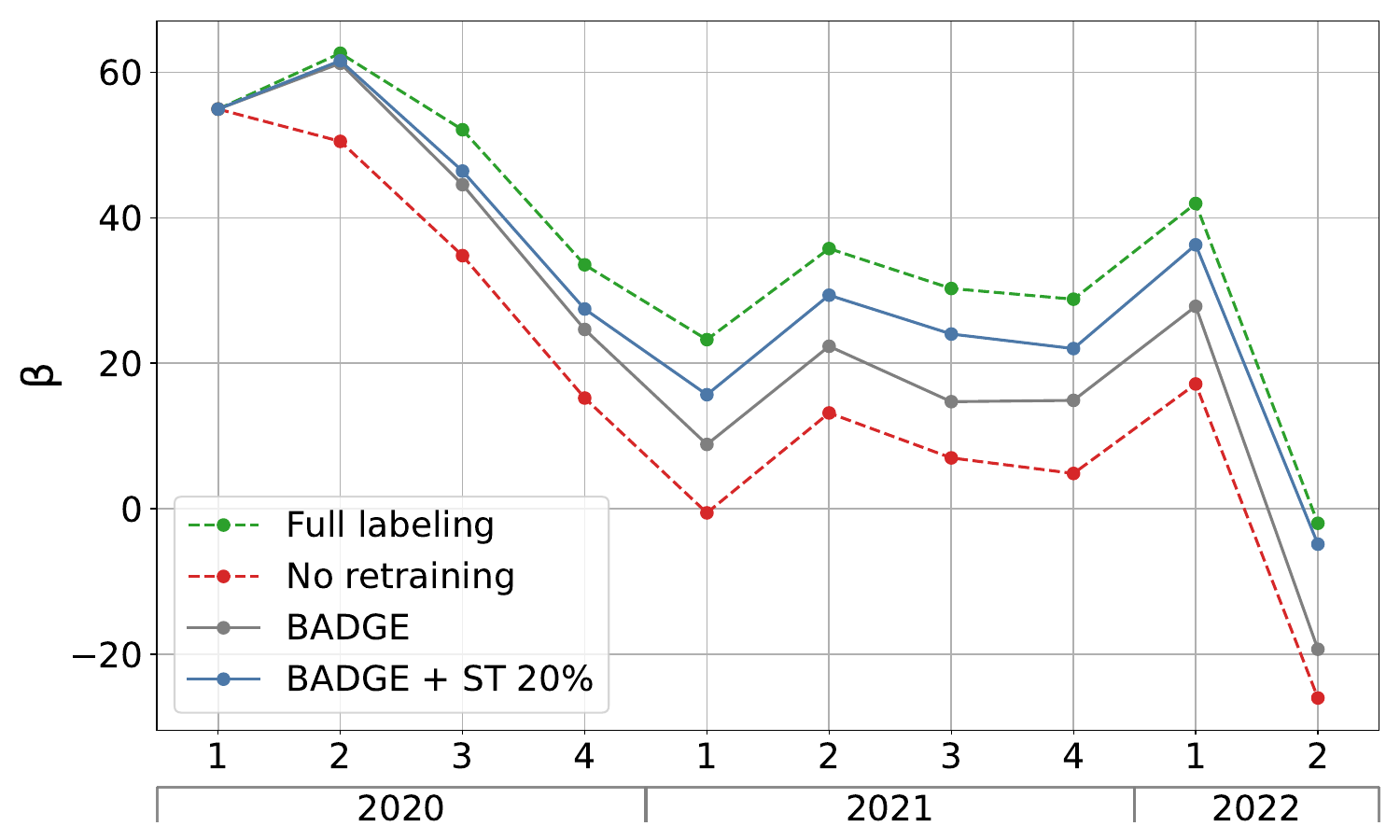}
            \label{fig:ELSA_active_features}
        }

        \hspace{-1em}
        \subfigure[EMBER – BADGE (Budget - 2\%)]{
            \includegraphics[trim=1.2cm 0cm 0cm 0cm, clip, width=0.46\textwidth]{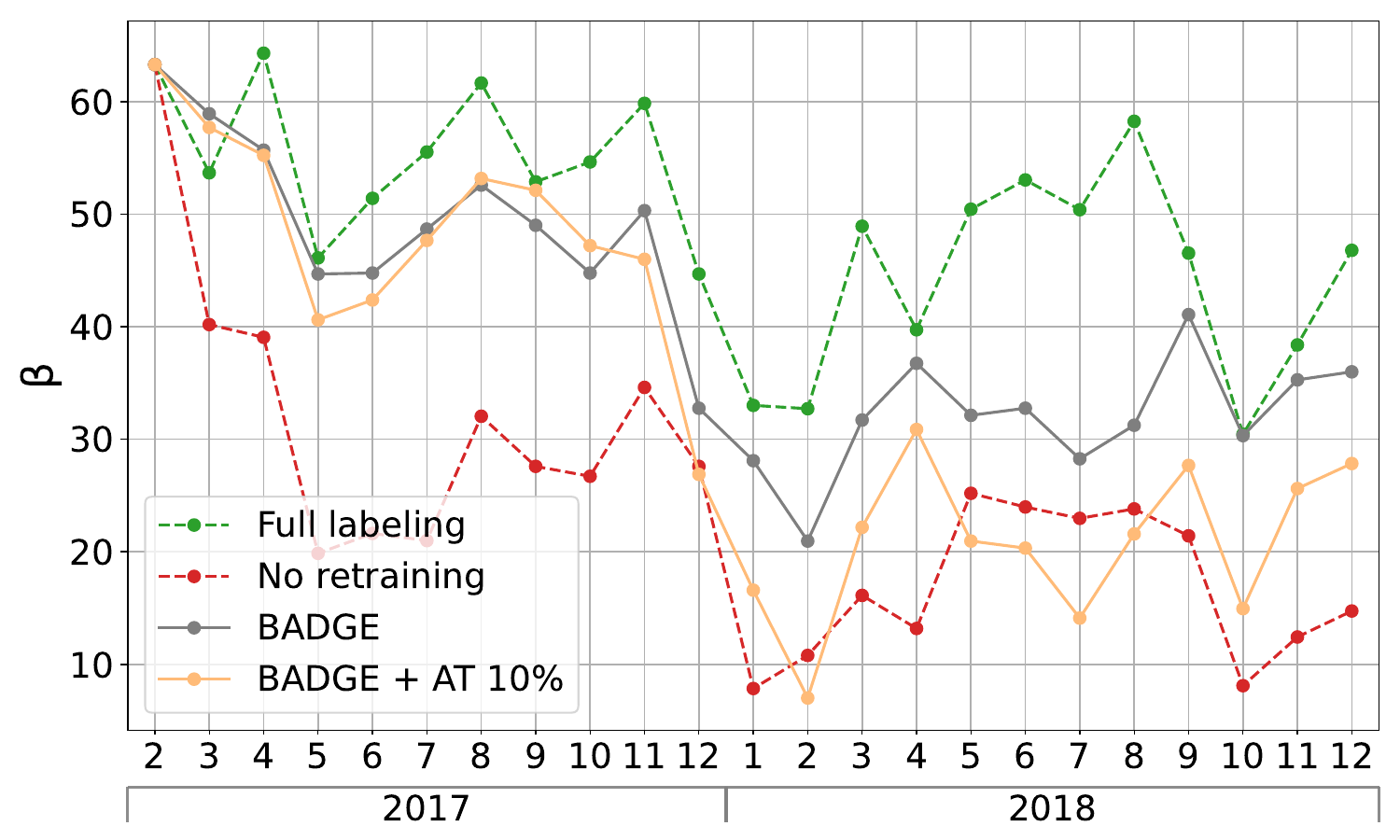}
            \label{fig:EMBER_active_features}
        }
        \hspace{-4em}

    }
    
    \caption{F1 score at FPR \(=1\%\) over time using RF on ELSA (quarterly) and EMBER (monthly), where each AL strategy is combined with its best-performing SSL strategy, including the optimal SSL budget, at a fixed labeling budget of \(1\%\) (ELSA) and \(2\%\) (EMBER). The lower panel reports the discriminant percentage of features over time, denoted by \(\beta\), corresponding to the performance curves above.}
    
    \label{fig:AL_and_SSL_f1_budgets}
\end{figure*}
\myparagraphdot{Combining AL and SSL}
We now analyze the impact of combining AL with SSL using the pipeline described in \autoref{Sect:AL_SSL_combined}. For each AL--SSL pairing, we consider multiple budgets for AL (1\%, 2\%, 5\%, 10\%) and SSL (10\%, 20\%, 40\%, 80\%), and compare each hybrid configuration against the corresponding AL-only baseline to isolate the contribution of pseudo-labeling after selective labeling. Due to space constraints, we do not report the entire resulting comparison grid, but summarize the observed dominant trends for representative low-budget configurations, where interaction between AL and SSL is most informative.
In particular, Figs.~\ref{fig:AL_SSL_ELSA_BADGE}~and~\ref{fig:AL_SSL_EMBER_BADGE} show the F1 score over time, with the AL budget fixed to $1\%$ on ELSA and $2\%$ on EMBER. For each dataset, we report only the SSL configuration that maximizes F1 at that budget, including both the selected SSL method and its pseudo-label budget. This enables a direct comparison between the best AL+SSL configuration and the corresponding AL-only baseline, thereby isolating SSL's contribution.
At low AL budgets, adding SSL yields clear gains. On ELSA, with AL budget fixed to $1\%$, the best-performing hybrid configuration combines BADGE with \textsc{ST} using a $20\%$ pseudo-label budget. On EMBER, with a to $2\%$ AL budget, the best-performing hybrid configuration combines BADGE with \textsc{AT} using a $10\%$ pseudo-label budget. In both cases, the hybrid strategy improves over the corresponding AL-only baseline and narrows the gap to FL, indicating that a small amount of newly acquired labels can make subsequent pseudo-labeling sufficiently reliable. Notably, the best results are obtained with relatively small pseudo-label budgets, suggesting that, in this low-budget regime, modest SSL updates are preferable to more aggressive pseudo-labeling, as they provide additional coverage while limiting error propagation.
As the AL budget increases, the benefits of SSL diminish, as AL alone already acquires enough informative labeled samples to effectively update the detector, approaching the FL baseline upper bound. In this regime, the marginal benefit of SSL depends on the configuration and may be offset by the noise introduced by incorrect pseudo-labels. This suggests that the main contribution of SSL is concentrated in the most label-constrained regimes, whereas with larger AL budgets, selective labeling alone is often sufficient to achieve near-optimal adaptation.
These results reveal a consistent pattern: SSL is most beneficial when the labeling budget is limited, reducing the amount of labeled data needed to approach a target performance. As the AL budget increases, the AL-only baseline approaches the performance of full-retraining, and SSL pseudo-labeling does not provide significant gains. 
%
\begin{insightbox}
    Combining AL and SSL is most effective in low-budget labeling regimes ($<2\%$).
\end{insightbox}

\subsection{Experimental Results on Distribution Drift}

\begin{figure}[t]
    \makebox[\textwidth][l]{%
        \hspace{-1em}
        \subfigure{
            \includegraphics[trim= 0cm 0cm 0cm 0cm, clip, width=0.24\textwidth]{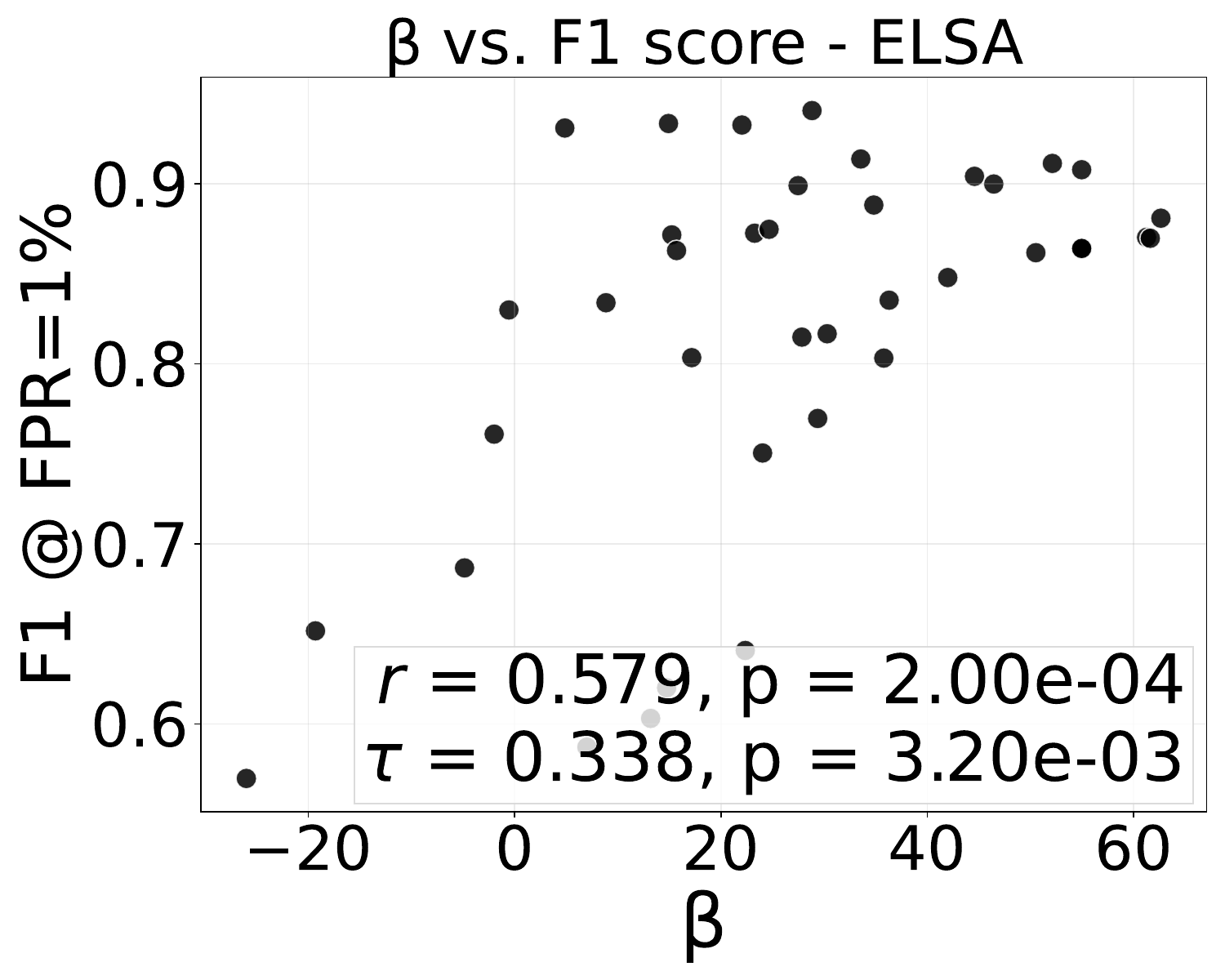}
            \label{fig:scatter_ELSA}
        }
        \subfigure{
            \includegraphics[trim=0cm 0cm 0cm 0cm, clip, width=0.24\textwidth]{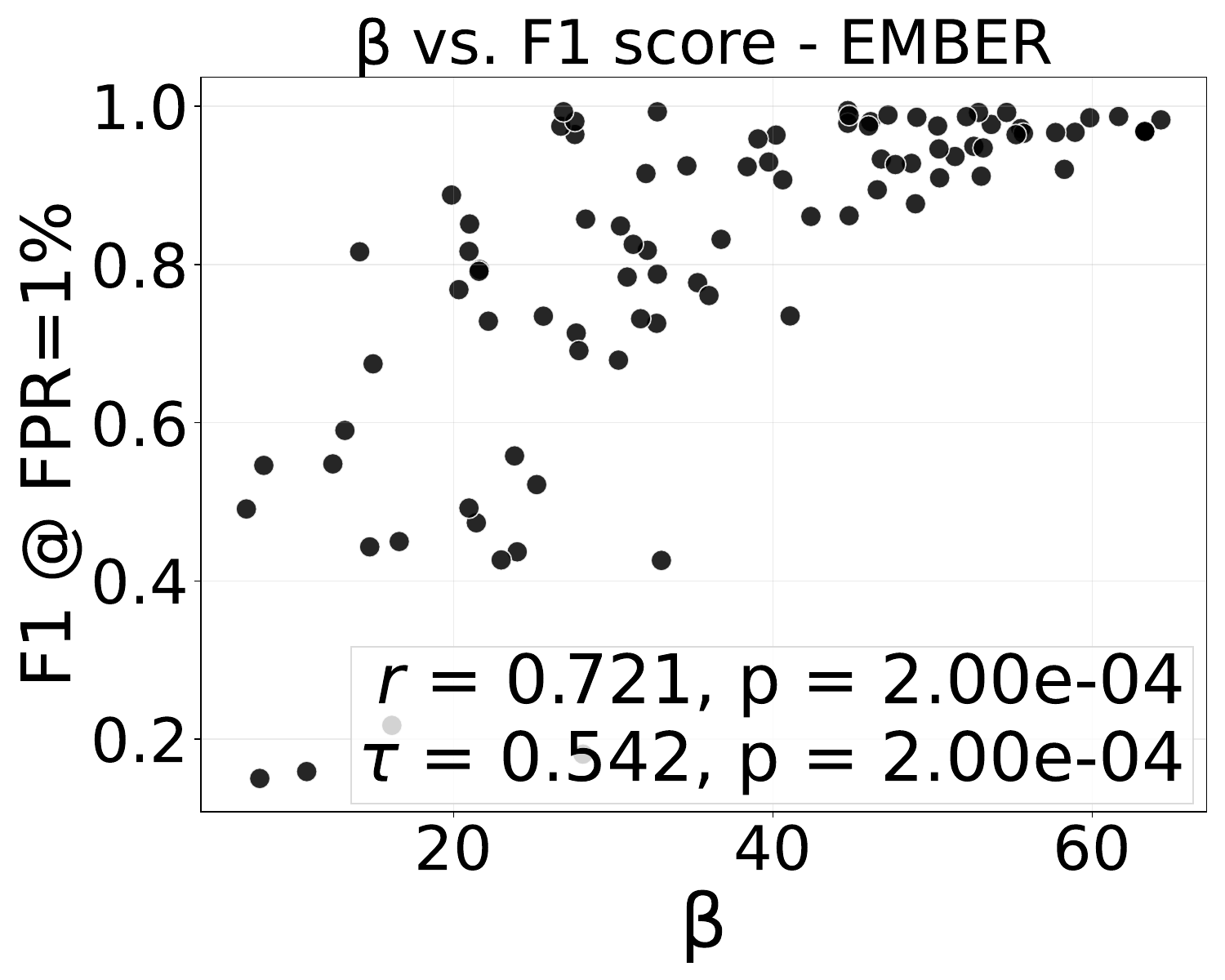}
            \label{fig:scatter_EMBER}
        }

        \hspace{-4em}
    }
    \caption{Scatter plots of F1 score versus $\beta$ of the curve in \autoref{fig:AL_and_SSL_f1_budgets} for ELSA (right plot) and EMBER (left plot). The insets report Pearson $r$ and Kendall $\tau$ correlations with two-sided permutation-test p-values (10,000 resamples).}
    \label{fig:scatter_both}
\end{figure}

\label{Sect:results_drift}
While previous results quantify the effectiveness of AL and SSL in terms of detection performance, they do not explain why some retraining strategies adapt better to distribution drift than others. We therefore complement the performance analysis with a feature-level drift diagnosis.
The datasets exhibit heterogeneous drift conditions: EMBER shows batch-level class-prior fluctuations and a discontinuity across the 2017/2018 release boundary, whereas ELSA is strongly imbalanced and exhibits time-varying feature activity patterns. To enable a unified diagnosis across these settings, we rely on the feature-level stability indicator introduced in \autoref{Sect:drift}.
In particular, at each time step $t$ we compute the stability score $\beta^{(t)}$, i.e., the percentage of features that preserve their signed class association (if statistically significant) minus the percentage of features that change association direction from the current training set to the subsequent evaluation batch. 
We then compare how each $\beta^{(t)}$ relates to the corresponding $\mathrm{F1}^{(t)}$ analyzing the same retraining policies shown in \autoref{fig:AL_and_SSL_f1_budgets}. Specifically, for both datasets we consider the two baselines (FL and NR), the best-performing AL configuration in each domain (BADGE at $1\%$ on ELSA and BADGE at $2\%$ on EMBER), and their best SSL-augmented variants (BADGE+\textsc{ST} with a $20\%$ SSL budget on ELSA and BADGE+\textsc{AT} with a $10\%$ SSL budget on EMBER). The corresponding stability scores $\beta^{(t)}$ are plotted in Figs.~\ref{fig:ELSA_active_features}~and~\ref{fig:EMBER_active_features}, directly beneath the associated F1 curves, so that temporal changes in feature stability can be visually compared with the corresponding performance dynamics. Results show a common pattern: temporal drops in $\beta^{(t)}$ are often accompanied by drops in the corresponding $\mathrm{F1}^{(t)}$ curves, and both tend to recover when adaptation is more effective.
NR generally exhibits lower stability and lower performance, whereas AL increases $\beta^{(t)}$ and AL+SSL often narrows the gap to FL, which remains the most stable and best-performing reference. The effect is more pronounced on ELSA, where both $\beta^{(t)}$ and F1 exhibit larger temporal swings, while EMBER shows smoother dynamics. Notably, on EMBER, BADGE+\textsc{AT} is in some batches slightly below BADGE alone, likely because AL already provides a strong update and the additional pseudo-labels introduce mild noise.
This visual co-variation suggests that retraining strategies that better preserve discriminative feature--class associations also tend to maintain higher detection performance over time.
To assess this effect more rigorously, we aggregate observations across retraining policies and compute the correlation between $\{\beta^{(t)}\}_{t=1}^{T}$ and $\{\mathrm{F1}^{(t)}\}_{t=1}^{T}$ using paired permutation tests for Pearson's $r$ and Kendall's $\tau$. This aggregation is particularly required for ELSA, which contains only 10 quarterly time points and would otherwise make per-policy tests underpowered. In this setting, we observe, in \autoref{fig:scatter_both} a statistically significant positive association in both datasets: on ELSA, $r=0.579$ ($p=2.0\times 10^{-4}$) and $\tau=0.338$ ($p=3.20\times 10^{-3}$); on EMBER, $r=0.721$ ($p=2.0\times 10^{-4}$) and $\tau=0.542$ ($p=2.0\times 10^{-4}$).
Overall, these results support the effectiveness of $\beta^{(t)}$ as a drift-sensitive diagnostic: higher stability is generally associated with better detection performance, although the relationship is not one-to-one. Indeed, configurations with similar $\beta^{(t)}$ can still yield different F1 scores, indicating that stability is informative, but not sufficient by itself to fully predict performance. This highlights the influence of the specific retraining policy, since performance depends not only on how much discriminative structure is refreshed, but also on which feature--class associations are updated.

\begin{insightbox}
   Higher stability of discriminant features over time is associated with better detection performance.
\end{insightbox}

\section{Related Work}\label{Sect:related}

\myparagraphdot{AL for Malware Detection} 
AL for malware detection under distribution drift remains a largely underexplored area. 
ActDroid~\cite{muzaffar2024actdroidactivelearningframework} is an online learning framework that employs a single uncertainty sampling strategy based on prediction confidence thresholds, specifically designed to detect Android malware. In contrast, the work presented in~\cite{chen2023continuous} proposes a technique that combines contrastive learning with a pseudo-loss uncertainty score to facilitate active sample selection in a continuous Android malware detection pipeline. 

Existing studies show AL’s potential for distribution drift but lack systematic comparisons across AL strategies and malware domains. Our framework addresses this gap with a unified evaluation using consistent budgets and models.

\myparagraphdot{SSL for Malware Detection}
The role of SSL in malware detection under distribution drift has received limited attention. DroidEvolver~\cite{xu2019droidevolver} employs an ensemble of classifiers incrementally updated via pseudo labeling. DroidEvolver++~\cite{kan2021investigating} improves robustness to distribution drift by refining pseudo labeling, stabilizing label assignments, and adapting detection thresholds. Both are limited to the Android domain. MORPH~\cite{alam2024morphautomatedconceptdrift}, instead, uses periodic self-training with asymmetric pseudo-labeling tailored for neural networks, while ADAPT~\cite{alam2025adapt} applies adaptive confidence-based self-training. 
Existing methods often yield unfair comparisons by evaluating SSL strategies under inconsistent model architectures and labeling budgets, leading to biased results. Our framework enables controlled comparisons within a unified setup, varying labeling budgets, and covering multiple malware domains.

\myparagraphdot{Combining AL and SSL} A small body of prior work has explored combining AL and SSL techniques in the malware domain. 
MalOSDF~\cite{guo2024malosdf} combines opcode-slice feature extraction with SSEAL, a semi-supervised ensemble that leverages an AL algorithm to reduce labeling cost by pseudo-labeling confident samples and querying uncertain ones for annotation. However, this approach is not evaluated under distribution drift and is limited to Windows malware, lacking comparison with alternative AL or SSL methodologies.
LDCDroid~\cite{liu2025ldcdroid} mitigates model aging in Android malware detection by selecting samples for retraining based on distribution drift measured in model-specific latent features. Nevertheless, it remains tied to a single neural architecture, it is limited to the Android domain, and it explores only a limited set of AL strategies.
In contrast, our work adopts a model-agnostic and modular perspective, enabling a systematic evaluation of standard AL and SSL strategies—both individually and in combination—under realistic temporal evolution. Moreover, we complement performance evaluation with a dedicated analysis of distribution drift to inspect how different retraining strategies update the model over time, an aspect largely unexplored in prior work.

\section{Conclusion and Future Work}\label{Sect:conclusions}
This work addresses the problem of efficiently updating malware classifiers while reducing reliance on costly annotations.
We introduce a model-agnostic retraining pipeline with modular components combining AL and SSL techniques. Across Android and Windows domains, we show that AL can approach the effectiveness of full labeling while requiring only $10\%$ of the labels, and that the proposed hybrid AL+SSL strategy further improves over AL alone by complementing sparse oracle queries with high-confidence pseudo-labels. 
We also introduce a feature-level analysis of distribution drift based on the stability of discriminative feature behavior over time. The proposed metric closely tracks detection performance over time and shows that effective retraining policies are those that better preserve (or refresh) discriminative features, narrowing the gap to full labeling; moreover, retraining strategies can differ in performance even at comparable stability, indicating that the techniques that select the best samples with the best features matter.
Our evaluation does not incorporate richer representation learning (e.g., transformer-based) or more advanced query strategies beyond the examined AL heuristics.
Future work will extend the framework to include such methods and investigate stronger AL and SSL schemes tailored to adversarially evolving data. A promising direction is to couple the retraining pipeline with adaptive, feature-stability policies that dynamically decide when to retrain and the labeling budget, improving automation and annotation efficiency of malware detectors in rapidly-changing threat environments.

\section*{Acknowledgements}
This research was partially supported by the Horizon Europe projects ELSA (GA no. 101070617) and CoEvolution (101168560), and by SERICS (PE00000014) and FAIR (PE00000013) under the MUR NRRP (EU-NGEU).
This work was conducted while D. Ghiani was enrolled in the Italian National Doctorate on AI run by the Sapienza Univ. of Rome in collaboration with the Univ. of Cagliari.

\bibliographystyle{ieeetr}
\bibliography{bibliography}

@inproceedings{arp2014drebin,
  title={Drebin: Effective and explainable detection of android malware in your pocket.},
  author={Arp, Daniel and Spreitzenbarth, Michael and Hubner, Malte and Gascon, Hugo and Rieck, Konrad and Siemens, CERT},
  booktitle={Ndss},
  year={2014}
}

@misc{anderson2018ember,
      title={EMBER: An Open Dataset for Training Static PE Malware Machine Learning Models}, 
      author={Hyrum S. Anderson and Phil Roth},
      year={2018},
      eprint={1804.04637},
      archivePrefix={arXiv},
      primaryClass={cs.CR}
}

@inproceedings{pendlebury2019tesseract,
  title={{TESSERACT}: Eliminating experimental bias in malware classification across space and time},
  author={Pendlebury, Feargus and Pierazzi, Fabio and Jordaney, Roberto and Kinder, Johannes and Cavallaro, Lorenzo},
  booktitle={28th {USENIX} Security Symposium},
  year={2019}
}

@inproceedings{Allix:2016:ACM:2901739.2903508,
    author = {Allix, Kevin and Bissyand{\'e}, Tegawend{\'e} F. and Klein, Jacques and Le Traon, Yves},
    title = {AndroZoo: Collecting Millions of Android Apps for the Research Community},
    booktitle = {Proceedings of the 13th Int'l Conf. on Mining Software Repositories},
    series = {MSR '16},
    year = {2016},
}

@techreport{settles2009active,
  title={Active Learning Literature Survey},
  author={Burr Settles},
  institution={University of Wisconsin–Madison},
  year={2009},
  url={https://api.semanticscholar.org/CorpusID:324600}
}

@inproceedings{wang2018uncertainty,
  title={Uncertainty sampling for action recognition via maximizing expected average precision},
  author={Wang, Hanmo and Chang, Xiaojun and Shi, Lei and Yang, Yi and Shen, Yi-Dong},
  booktitle={IJCAI international joint conference on artificial intelligence},
  year={2018}
}

@inproceedings{prabhu2021activedomainadaptationclustering,
  title={Active Domain Adaptation via Clustering Uncertainty-weighted Embeddings},
  author={Prabhu, Viraj and Chandrasekaran, Arjun and Saenko, Kate and Hoffman, Judy},
  booktitle={2021 IEEE/CVF International Conference on Computer Vision},
  year={2021},
}

@inproceedings{chen2023continuous,
  title={Continuous learning for android malware detection},
  author={Chen, Yizheng and Ding, Zhoujie and Wagner, David},
  booktitle={32nd USENIX Security Symposium},
  year={2023}
}

@article{guo2024malosdf,
  title={MalOSDF: An Opcode Slice-Based Malware Detection Framework Using Active and Ensemble Learning},
  author={Guo, Wenjie and Xue, Jingfeng and Meng, Wenheng and Han, Weijie and Liu, Zishu and Wang, Yong and Li, Zhongjun},
  journal={Electronics},
  year={2024},
}

@misc{alam2024morphautomatedconceptdrift,
      title={MORPH: Towards Automated Concept Drift Adaptation for Malware Detection}, 
      author={Md Tanvirul Alam and Romy Fieblinger and Ashim Mahara and Nidhi Rastogi},
      year={2024},
      eprint={2401.12790},
      archivePrefix={arXiv},
      primaryClass={cs.LG},
      url={https://arxiv.org/abs/2401.12790}, 
}

@article{muzaffar2024actdroidactivelearningframework,
  title={ActDroid: An active learning framework for Android malware detection},
  author={Muzaffar, Ali and Hassen, Hani Ragab and Zantout, Hind and Lones, Michael A},
  journal={Computers \& Security},
  year={2025},
}

@inproceedings{xu2019droidevolver,
  title={Droidevolver: Self-evolving android malware detection system},
  author={Xu, Ke and Li, Yingjiu and Deng, Robert and Chen, Kai and Xu, Jiayun},
  booktitle={2019 IEEE European Symposium on Security and Privacy (EuroS\&P)},
  year={2019},
}

@article{vanEngelen2020,
  author    = {van Engelen, J. E. and Hoos, H. H.},
  title     = {A Survey on Semi-Supervised Learning},
  journal   = {Machine Learning},
  year      = {2020},
}

@inproceedings{ash2020badge,
  author       = {Jordan T. Ash and
                  Chicheng Zhang and
                  Akshay Krishnamurthy and
                  John Langford and
                  Alekh Agarwal},
  title        = {Deep Batch Active Learning by Diverse, Uncertain Gradient Lower Bounds},
  booktitle    = {8th Int'l Conf. Learn. Repr. ({ICLR})},
  year         = {2020},
}

@inproceedings{sener2017active,
  title={Active Learning for Convolutional Neural Networks: A Core-Set Approach},
  author={Sener, Ozan and Savarese, Silvio},
  booktitle={Int'l Conf. Learn. Repr. (ICLR)},
  year={2018}
}

@inproceedings{kan2021investigating,
  title={Investigating labelless drift adaptation for malware detection},
  author={Kan, Zeliang and Pendlebury, Feargus and Pierazzi, Fabio and Cavallaro, Lorenzo},
  booktitle={Proc. 14th ACM Workshop on AI and Security ({AISec})},
  year={2021}
}

@article{alam2025adapt,
  title={ADAPT: A Pseudo-labeling Approach to Combat Concept Drift in Malware Detection},
  author={Alam, Md Tanvirul and Piplai, Aritran and Rastogi, Nidhi},
  journal={arXiv preprint arXiv:2507.08597},
  year={2025}
}

@article{liu2025ldcdroid,
  title={LDCDroid: Learning data drift characteristics for handling the model aging problem in Android malware detection},
  author={Liu, Zhen and Wang, Ruoyu and Peng, Bitao and Qiu, Lingyu and Gan, Qingqing and Wang, Changji and Zhang, Wenbin},
  journal={Computers \& Security},
  year={2025},
}

@article{joyce2023MOTIF,
title = {MOTIF: A Malware Reference Dataset with Ground Truth Family Labels},
journal = {Computers \& Security},
year = {2023},
author = {Robert J. Joyce and Dev Amlani and Charles Nicholas and Edward Raff},
}

@article{MANIRIHO2024windowsMLSurvey,
title = {A systematic literature review on Windows malware detection: Techniques, research issues, and future directions},
journal = {Journal of Systems and Software},
year = {2024},
author = {Pascal Maniriho and Abdun Naser Mahmood and Mohammad Jabed Morshed Chowdhury},
}

@article{MUZAFFAR2022androidMLSurvey,
title = {An in-depth review of machine learning based Android malware detection},
journal = {Computers \& Security},
year = {2022},
author = {Ali Muzaffar and Hani {Ragab Hassen} and Michael A. Lones and Hind Zantout},
}

@inproceedings{stanescu2014asymmetric,
  title={Semi-supervised self-training approaches for imbalanced splice site datasets},
  author={Stanescu, Ana and Caragea, Doina},
  booktitle={Proc. of the 6th Int'l Conference on Bioinformatics and Computational Biology, BICoB},
  year={2014}
}

@inproceedings{yarowsky1995unsupervised,
  title={Unsupervised word sense disambiguation rivaling supervised methods},
  author={Yarowsky, David},
  booktitle={33rd annual meeting of the association for computational linguistics},
  year={1995}
}

@article{chen2022debiased,
  title={Debiased self-training for semi-supervised learning},
  author={Chen, Baixu and Jiang, Junguang and Wang, Ximei and Wan, Pengfei and Wang, Jianmin and Long, Mingsheng},
  journal={Advances in Neural Information Processing Systems},
  year={2022}
}

@misc{ramd,
    title = {Robust Android Malware Detection Competition},
    note = {Accessed on May 2025},
    howpublished = {\url{https://ramd-competition.github.io/}}
}

@inproceedings{minnei2025experimental,
  title={An Experimental Analysis of Semi-supervised Learning for Malware Detection},
  author={Minnei, Luca and Piras, Giorgio and Sotgiu, Angelo and Pintor, Maura and Demontis, Ambra and Maiorca, Davide and Biggio, Battista and others},
  booktitle={CEUR WORKSHOP PROCEEDINGS},
  year={2025}
}

%
\vskip -1\baselineskip plus -0fil
\begin{IEEEbiography}[{\includegraphics[width=1in,height=1.25in,clip,keepaspectratio]{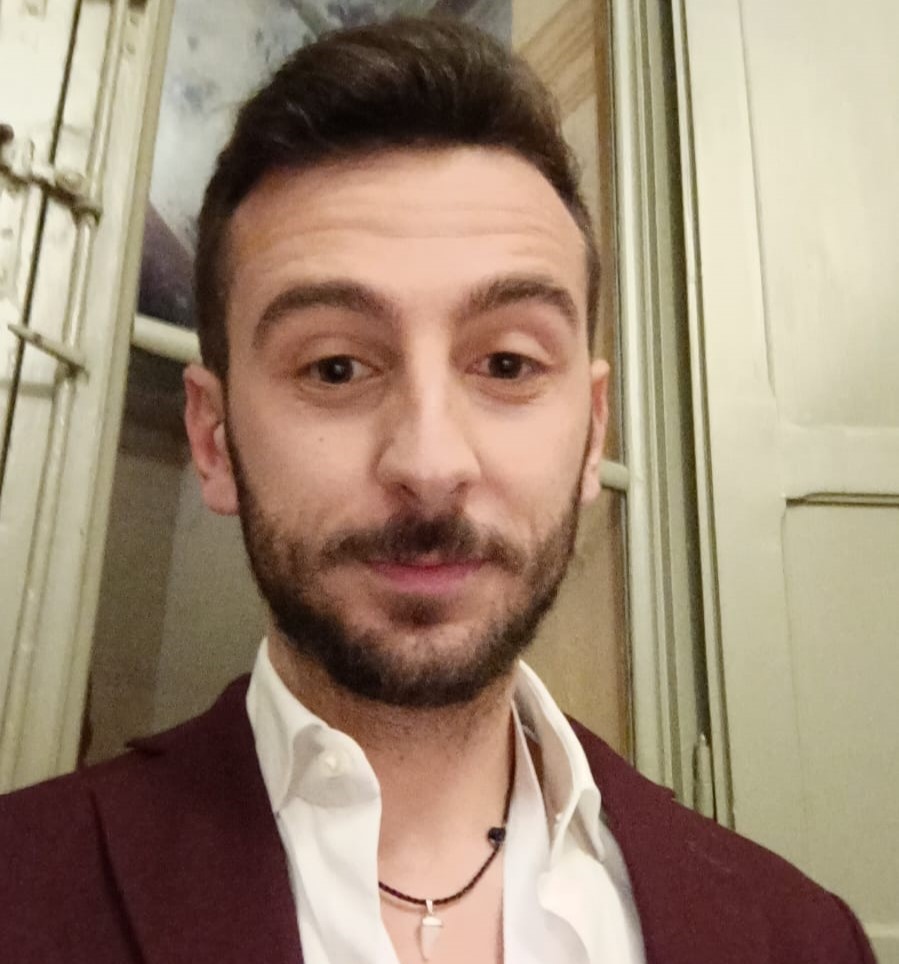}}]{Luca Minnei} is a Ph.D. student in Informatics, Electronics, and Computer Engineering at the University of Cagliari, Italy, he earned a B.Sc. in Computer Science in 2022 and an M.Sc. in Computer Engineering, Cybersecurity, and Artificial Intelligence in 2024, both with honors. His research focuses on malware detection and concept drift in machine learning security.
\end{IEEEbiography}
\vskip -1\baselineskip plus -0fil
\begin{IEEEbiography}[{\includegraphics[width=1in,height=1.25in,clip,keepaspectratio]{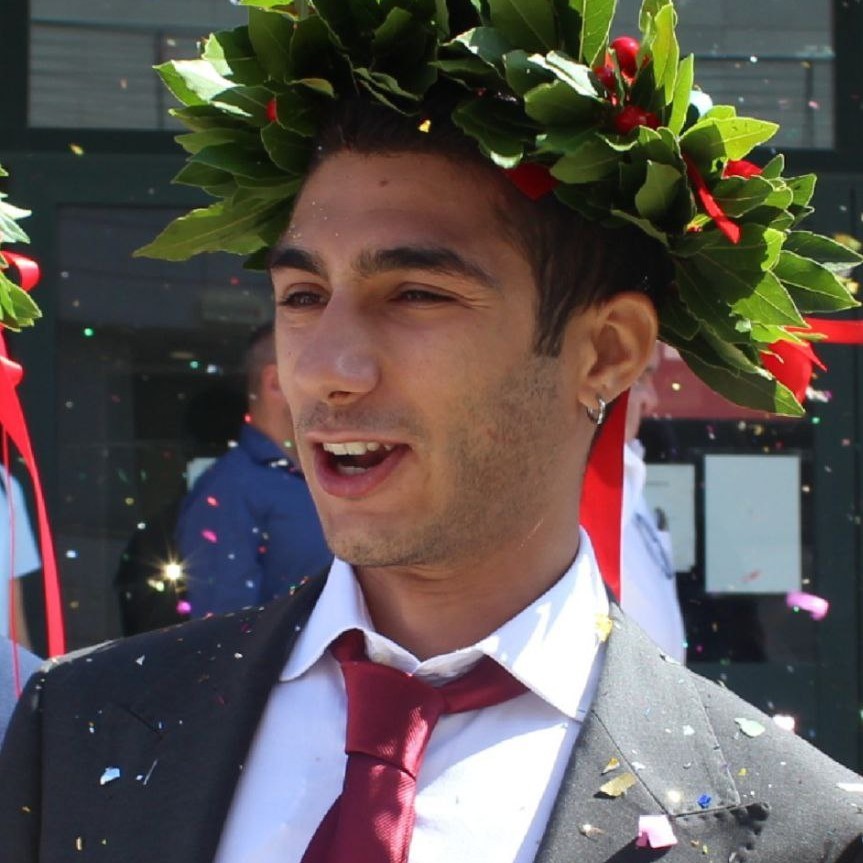}}]{Cristian Manca} is a Master's student in Computer Engineering, Cybersecurity, and Artificial Intelligence. He earned his B.Sc. in Electrical, Electronic, and Computer Engineering in 2023, with a thesis on the identification and rejection of evasive samples with Neural Rejection Defense.
\end{IEEEbiography}
\vskip -1\baselineskip plus -0fil
\begin{IEEEbiography}[{\includegraphics[width=1in,height=1.25in,clip,keepaspectratio]{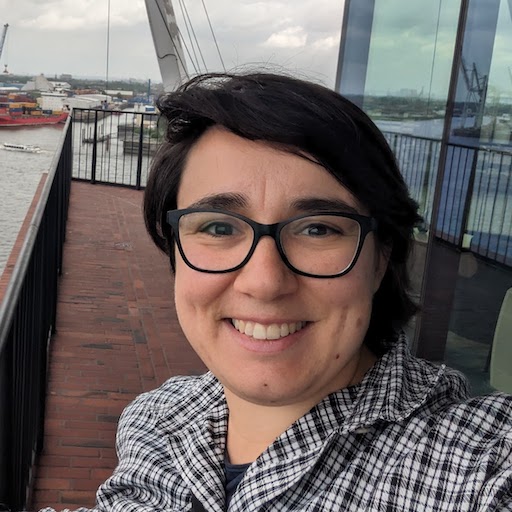}}]{Maura Pintor}is an Assistant Professor at the University of Cagliari, Italy. She received her PhD in Electronic and Computer Engineering (with honors) in 2022 from the University of Cagliari. Her research interests include adversarial machine learning and trustworthy security evaluations of ML models, with applications in cybersecurity. 
She serves as an AC for NeurIPS, and as AE for Pattern Recognition.
\end{IEEEbiography}
\vskip -1\baselineskip plus -0fil
\begin{IEEEbiography}[{\includegraphics[width=1in,height=1.25in,clip,keepaspectratio]{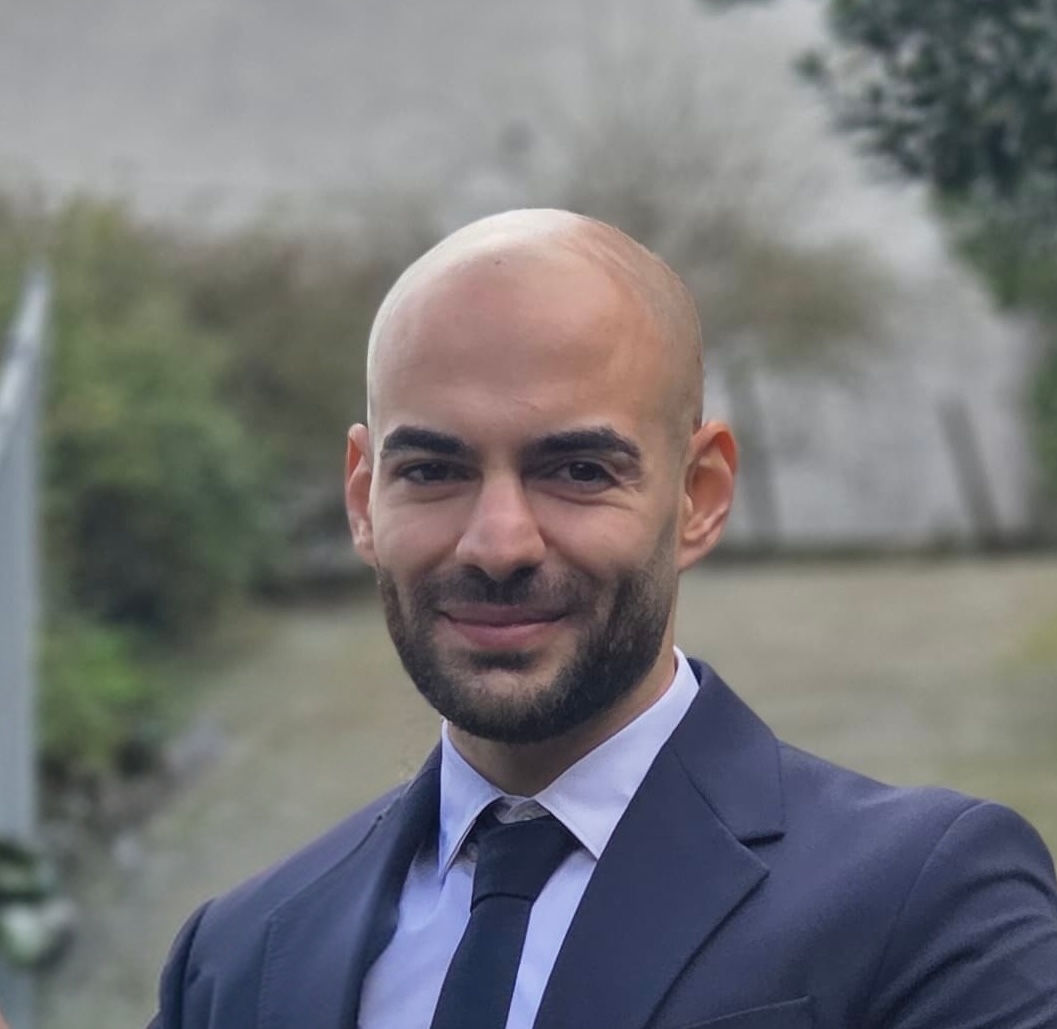}}]{Giorgio Piras} is a Postdoctoral Researcher at the University of Cagliari. He received his PhD in Artificial Intelligence in January 2025 from the Sapienza University of Rome (with honors). His research mainly focuses on adversarial machine learning, with a particular attention to neural network pruning, explainable AI, and LLM security. He serves as a reviewer for journals and conferences, including Pattern Recognition and Neurocomputing journals.
\end{IEEEbiography}
\vskip -1\baselineskip plus -0fil
\begin{IEEEbiography}[{\includegraphics[width=1in,height=1.25in,clip,keepaspectratio]{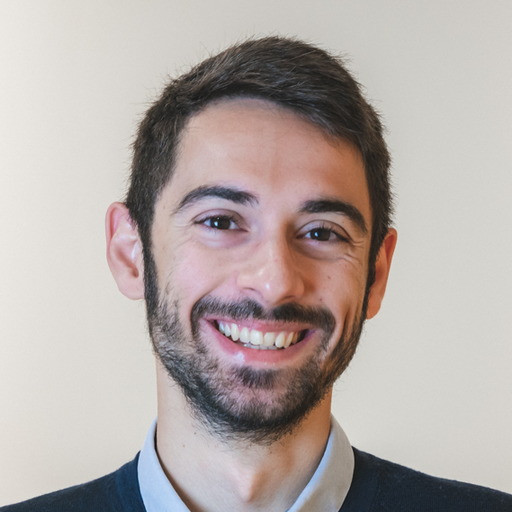}}]{Angelo Sotgiu} is an Assistant Professor at the University of Cagliari. He received from the University of Cagliari (Italy) the PhD in Electronic and Computer Engineering in February 2023. His research mainly focuses on the security of machine learning, also considering practical applications like malware detection. He serves as a reviewer for several journals and conferences.
\end{IEEEbiography}
\vskip -1\baselineskip plus -0fil
\begin{IEEEbiography}[{\includegraphics[width=1in,height=1.25in,clip,keepaspectratio]{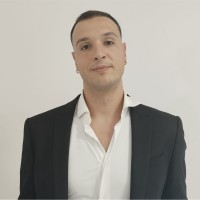}}]{Daniele Ghiani} received his BSc in Computer Engineering (2021) and MSc in Computer Engineering, Cybersecurity, and AI (2024) from the University of Cagliari. He is currently a PhD student in the Italian national PhD programme in AI at Sapienza University, co-located at the University of Cagliari. His research addresses Continual Learning and regression issues in Android malware detection.
\end{IEEEbiography}
\vskip -1\baselineskip plus -0fil
\begin{IEEEbiography}[{\includegraphics[width=1in,height=1.25in,clip,keepaspectratio]{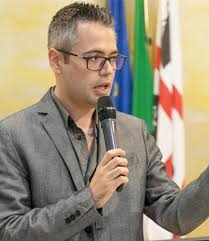}}]{Davide Maiorca} received the Ph.D. in 2016 from the University of Cagliari. He is an Associate Professor of Computer Engineering at the University of Cagliari and a member of the PRA Lab. His research focuses on x86/Android malware analysis and detection, malicious documents (e.g., PDF, Microsoft Office), and adversarial machine learning. He earned the Italian National Scientific Qualification (ASN) in 2021 and has authored 25+ papers.
\vspace{3em}
\end{IEEEbiography}
\vskip -1\baselineskip plus -0fil
\begin{IEEEbiography}[{\includegraphics[width=1in,height=1.25in,clip,keepaspectratio]{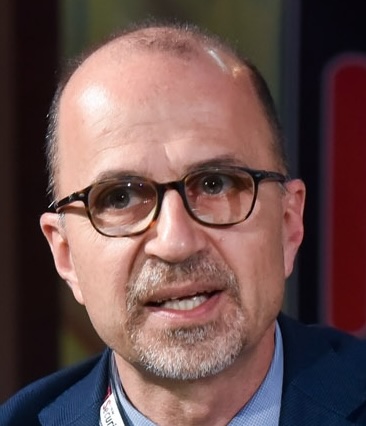}}]{Giorgio Giacinto} is a Professor of Computer Engineering at the University of Cagliari, Italy. He leads Cybersecurity research within the sAIfer Lab research group. His main contributions lie in machine learning approaches to cybersecurity, including threat analysis and detection, and are supported by funding from national and international projects. He has published nearly 200 papers in international conferences and journals and regularly serves as a member of the Editorial Board and Program Committee for several of them. He is a Fellow of the IAPR and a Senior Member of the IEEE Computer Society and ACM.
\end{IEEEbiography}

\vskip -1\baselineskip plus -0fil

\begin{IEEEbiography}[{\includegraphics[width=1in,height=1.25in,clip,keepaspectratio]{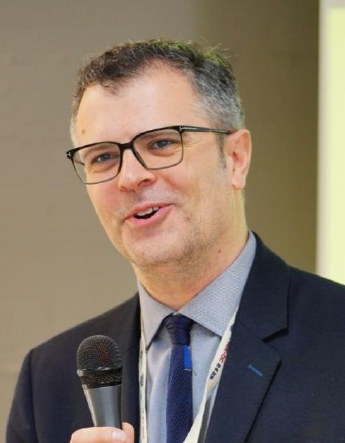}}]{Battista Biggio} (MSc 2006, PhD 2010) is Full Professor of Computer Engineering at the University of Cagliari, Italy. He has provided pioneering contributions to machine learning security. His paper ``Poisoning Attacks against Support Vector Machines'' won the prestigious 2022 ICML Test of Time Award.  He chaired IAPR TC1 (2016-2020), and served as Associate Editor for IEEE TNNLS and IEEE CIM. He is now Associate Editor-in-Chief for Pattern Recognition and serves as Area Chair for NeurIPS and IEEE Symp. SP. He is Fellow of IEEE and AAIA, ACM Senior Member, and Member of IAPR, AAAI, and ELLIS. 
\end{IEEEbiography}







\end{document}